\documentclass[10pt,twocolumn,letterpaper]{article}

\usepackage{iccv}              

\definecolor{iccvblue}{rgb}{0.21,0.49,0.74}
\usepackage[pagebackref,breaklinks,colorlinks,allcolors=iccvblue]{hyperref}
\usepackage{colortbl}
\usepackage{graphicx}
\usepackage{bbm}
\usepackage{multirow}
\usepackage{subcaption}
\usepackage{arydshln}
\usepackage{diagbox}

\usepackage[accsupp]{axessibility}  

\def\paperID{10690} 
\def\confName{ICCV}
\def\confYear{2025}

\title{Adaptive Routing of Text-to-Image Generation Requests Between\\ Large Cloud Model and Light-Weight Edge Model}

\author{Zewei Xin, Qinya Li\thanks{Q. Li and C. Niu are the corresponding authors.}, Chaoyue Niu\footnotemark[1], Fan Wu, Guihai Chen\\
Shanghai Jiao Tong University, China\\
{\tt\small \{xin\_zewei\_12138, qinyali, rvince\}@sjtu.edu.cn, \{fwu, gchen\}@cs.sjtu.edu.cn}
}

\begin{document}
\maketitle
\begin{abstract}
Large text-to-image models demonstrate impressive generation capabilities; however, their substantial size necessitates expensive cloud servers for deployment. Conversely, light-weight models can be deployed on edge devices at lower cost but often with inferior generation quality for complex user prompts. To strike a balance between performance and cost, we propose a routing framework, called \texttt{RouteT2I}, which dynamically selects either the large cloud model or the light-weight edge model for each user prompt. Since generated image quality is challenging to measure and compare directly, \texttt{RouteT2I} establishes multi-dimensional quality metrics, particularly, by evaluating the similarity between the generated images and both positive and negative texts that describe each specific quality metric. \texttt{RouteT2I} then predicts the expected quality of the generated images by identifying key tokens in the prompt and comparing their impact on the quality. \texttt{RouteT2I} further introduces the Pareto relative superiority to compare the multi-metric quality of the generated images. Based on this comparison and predefined cost constraints, \texttt{RouteT2I} allocates prompts to either the edge or the cloud. Evaluation reveals that \texttt{RouteT2I} significantly reduces the number of requesting large cloud model while maintaining high-quality image generation.
\end{abstract}    
\section{Introduction}


Nowadays, text-to-image (T2I) models like Imagen~\cite{baldridge2024imagen}, Stable Diffusion~\cite{rombach2022high}, and DALL·E~\cite{ramesh2021zero} have achieved significant success in generating diverse, high-quality images given user prompts. However, the impressive generation quality comes with large model and high cost. For example, Stable Diffusion 3.5~\cite{huggingfaceStabilityaistablediffusion35largeHugging} has 8 billion parameters. Such a large model necessitates reliance on cloud servers, leading to high serving cost. As shown in \cref{tab:t2i}, it is particularly costly in commercial scenarios with millions of requests. 

\begin{table}[!t]
\centering
\small
\renewcommand{\arraystretch}{0.9}
\begin{tabular}{@{}l|rr@{}}
\toprule
Text-to-Image Model                 & \#Param   & Pricing (\$/M)    \\ \hline
Stable Diffusion 1.6~\cite{rombach2022high}      & 0.86 B      & 9 K        \\
Stable Diffusion XL~\cite{podell2023sdxl}       & 2.6 B       & 9 K        \\
Stable Diffusion 3 Medium~\cite{esser2024scaling} & 2 B         & 35 K       \\
Stable Diffusion 3~\cite{esser2024scaling}        & 8 B         & 65 K       \\
Stable Diffusion 3.5~\cite{huggingfaceStabilityaistablediffusion35largeHugging}      & 8 B         & 65 K       \\ \bottomrule
\end{tabular}
\caption{For the Stable Diffusion models, newer versions come with more parameters and higher API costs.}
\label{tab:t2i}
\end{table}

The high cost of cloud-based image generation drives the trend of deploying T2I models on edge devices. Techniques such as quantization~\cite{li2023q,zhao2024mixdq,so2024temporal}, structural pruning~\cite{li2024snapfusion,Castells_2024_CVPR}, and reducing denoising steps~\cite{luo2023latent,liu2023instaflow,sauer2025adversarial} are used to minimize model sizes and speed up inference, showing that edge deployment is feasible. Compared to cloud models, edge models leverage users' local computing devices to provide instant services anytime and anywhere without cloud serving cost and communication overhead. However, the light-weight edge T2I models often have lower generation quality compared to the cloud large models.

\begin{figure}[!t]
 \centering
 \includegraphics[width=\linewidth]{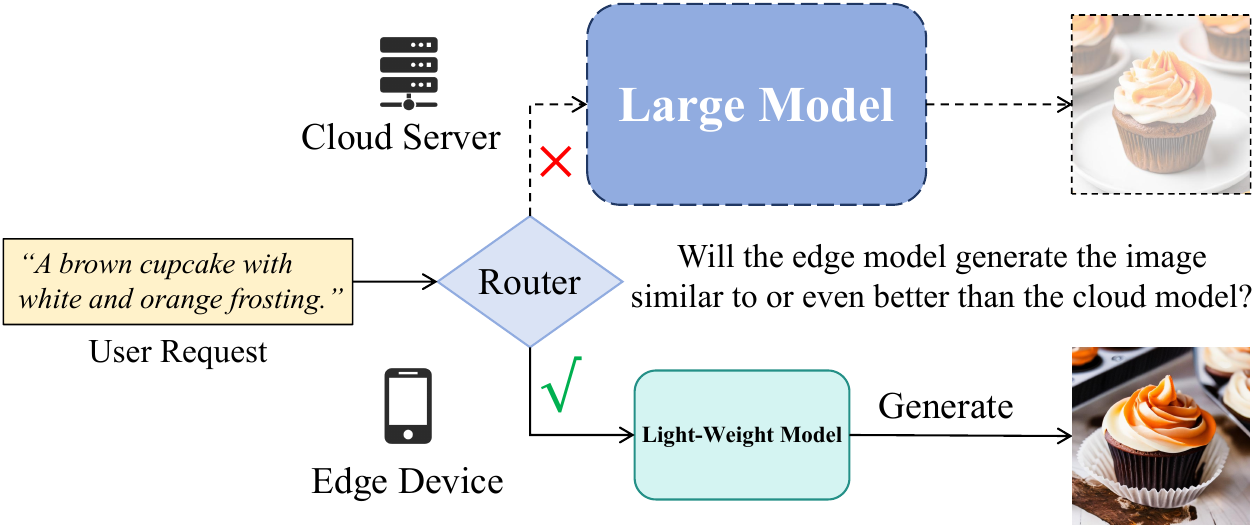}
 \caption{Routing for text-to-image generation requests.}
 \label{Route}
 \end{figure}

Although cloud models often offer superior quality, not all user prompts need these large models. For some prompts, light-weight edge models can produce comparable or even better results. To ensure high-image quality generation while limiting requests to cloud models to reduce cost, it is necessary to introduce a routing mechanism such that suitable models are selected for different user prompts based on their complexity. Intuitively, as shown in \cref{Route}, a routing framework for T2I models routes only challenging prompts to the cloud, while handling easy ones on the edge.

Previous routing methods are primarily designed for large language models (LLMs)~\cite{lu2023routing,ong2024routellm,ding2024hybrid,hu2024routerbench}, presenting significant challenges when applied to T2I models. One major challenge lies in comparing the quality of generated images. Unlike text, image quality does not have unified metrics and is subject to many influencing factors, including color, clarity, and the unique distortions typical of generated images~\cite{lee2024holistic}. This makes it hard to compare image quality to guide routing decisions. Furthermore, the output space of T2I models is substantially larger than the input text space, and a single prompt can correspond to numerous images, hindering the prediction of image quality before generation.

To address these challenges, we propose an edge-cloud T2I routing framework \texttt{RouteT2I}, deploying a routing model and a small T2I model on the edge, and a large T2I model on the cloud. Our goal is to optimize the overall quality of generated images at a limited cost, by effectively routing user prompts to the edge or the cloud. To evaluate the quality gap between generated images, we first propose a multi-metric quality measure, including metrics for both real photos and those unique to generated images. Each metric assesses image quality by measuring the similarity between the image and a pair of positive and negative prompts that describe the quality metric. We then define Pareto relative superiority (PRS) to quantify quality differences between images. \texttt{RouteT2I} predicts PRS based on user prompts, inspired by the cross-attention operation between text and image during generation. In particular, inspired by mixture-of-experts (MoE) selective focus on experts or tokens in LLMs~\cite{shazeer2017outrageously,zhou2022mixture,antoniak2025mixture}, we design a novel dual-gate token selection MoE, where a user prompt is treated as a sequence of tokens, with experts aligned to quality metrics. A token selection gate allows experts to proactively choose tokens that markedly impact these metrics, thereby focusing on key tokens. A dual-gate MoE uses both positive and negative gates to evaluate dominant effects when tokens have opposing effects simultaneously. Furthermore, multiple heads are introduced to predict multiple quality metrics. Based on the multi-metric quality prediction, \texttt{RouteT2I} determines quality differences between images generated on the edge over those in the cloud. Only user prompts that show substantial quality improvement when generated in the cloud are routed there, while others remain at the edge. Our key contributions are as follows:

\begin{itemize}
    \item We, for the first time, study how to route text-to-image generation requests between cloud and edge models for low cost and high quality.
    

    \item Specified to multi-metric quality measures of generated images, we design a routing model architecture with dual gate token selection MoE, combined with a routing strategy for user prompts.


    \item We evaluate the proposed design using 18 image generation model pairs on the public COCO 2014 dataset. Key results include: (1) by routing 50\% of requests to edge and cloud models, respectively, the overall quality improvement over the edge model reaches 83.97\% of that achieved by fully using the cloud model; and (2) to achieve 50\% of the quality improvement attained by fully using the cloud model, the number of requests routed to the cloud model is reduced by 70.24\% compared to a random routing policy.
    
\end{itemize}

\section{Related Work}

Model routing selects suitable models for user requests to ensure quality while reducing high cost of invoking powerful models. Depending on whether routing decision occurs before or after weak model execution, model routing can be categorized into predictive and non-predictive routing.

Non-predictive routing takes the output of a weak model to decide whether to route to the next, powerful model. LAECIPS~\cite{hu2024laecips} evaluated the confidence of predictions made by the edge semantic segmentation model to decide if cloud assistance from a large visual model is needed. For classification models, Hybrid~\cite{kag2023efficient} used a routing model to partition the data domain between edge and cloud models, routing cases with incorrect edge outputs to the cloud. For cascading LLM, LLM Cascade~\cite{yue2023large} routed based on the consistency of responses from weaker LLMs, while FrugalGPT~\cite{chen2023frugalgpt} considered current quality and past costs to determine routing. Tabi~\cite{wang2023tabi} not only routed based on the confidence of responses from weaker LLMs but also enhanced output quality by aggregating historic outputs.


Different from non-predictive routing, predictive routing selects model before running the weak model on resource-constrained edge device. Previous studies focused on LLM, with the difference in routing strategies. RouteLLM~\cite{ong2024routellm} predicted the outcomes of quality comparisons between LLM and routed based on prediction confidence. Hybrid LLM~\cite{ding2024hybrid} relaxed comparisons, allowing the weak model to succeed if the quality gap is within a threshold, thus saving costs with some quality compromise. Recent research~\cite{shnitzer2023large} advances the confidence routing strategy by identifying out-of-distribution data through confidence thresholds. ZOOTER~\cite{lu2023routing} introduced a new strategy by predicting the normalized quality of candidate model outputs, making routing decision based on relative quality, with routing models distilled from existing quality scoring models.


Previous routing work were mainly for LLMs, whereas the predictive routing for T2I in our work remains unexplored. A key distinction lies in the image outputs of T2I, which necessitate predicting quality variations in a larger output image space from a smaller input text space (Sec. B in Appendix). On one hand, unlike many LLM tasks with definitive answers, image quality lacks fixed standards and is highly subjective. Previous single-objective LLM routing struggles to adapt to such ambiguous quality criteria. On the other hand, prior work lacked dedicated routing model designs, making it challenging to predict quality relationships within the vast output image space. This underscores the necessity of effective routing model design.

\section{Problem Formulation}

In text-to-image generation scenario, the cloud normally hosts a large T2I model, denoted as $\mathcal{M}_c$, while a resource-constraint edge device, such as a smartphone, often deploys a light-weight T2I model, denoted as $\mathcal{M}_e$. For example, Google has released Imagen~\cite{saharia2022photorealistic,baldridge2024imagen} for cloud deployment and MobileDiffusion~\cite{zhao2023mobilediffusion} for edge deployment. From the perspective of a user, the large model generally offers superior generation quality, but incurs edge-cloud communication cost and cloud serving fee, which together are denoted as $F_c$. In contrast, the edge model can leverage the user's local device, thereby eliminating these additional costs.

\begin{figure}[!t]
\centering
\includegraphics[width=\linewidth]{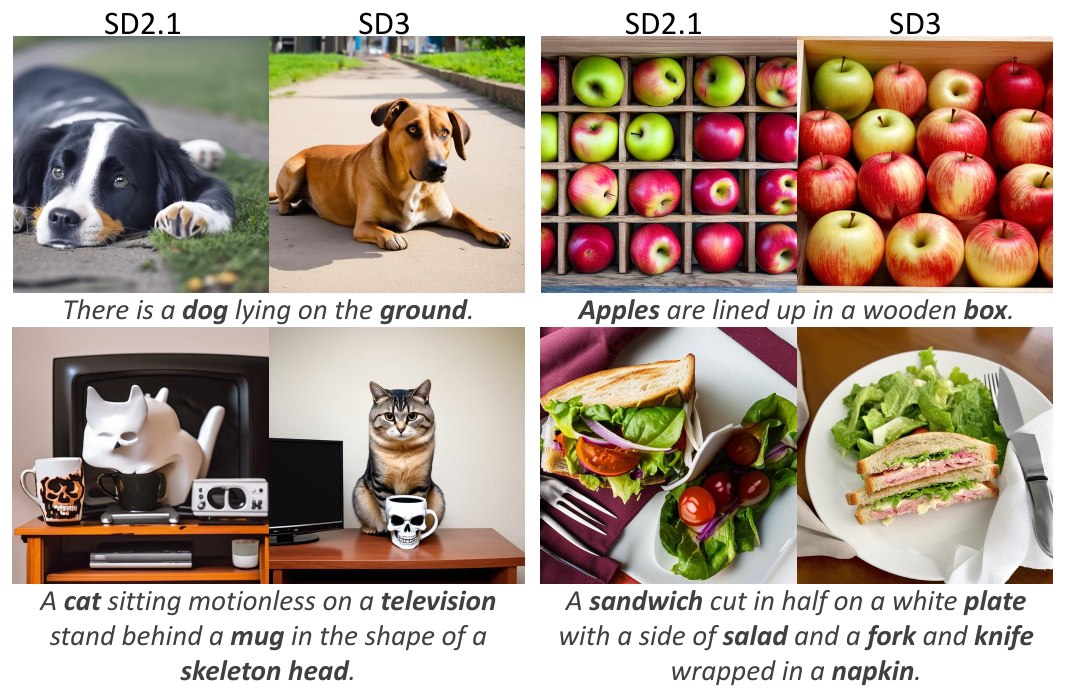}
\caption{SD2.1 tends to match or outperform SD3 on simple prompts, while more likely falling short on complex ones. Entities to be generated are \textbf{bolded}.}
\label{complex-example}
\end{figure}

\begin{figure}
    \centering
    \subcaptionbox{}[0.49\linewidth]{\includegraphics[width=\linewidth]{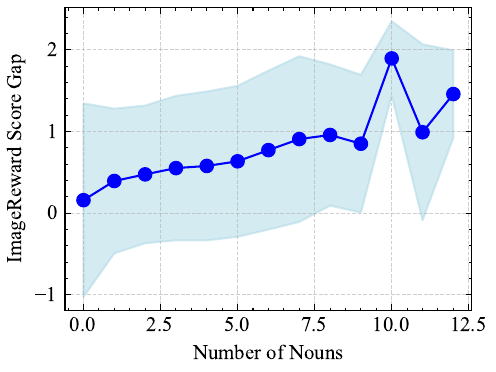}}
    \subcaptionbox{}[0.49\linewidth]{\includegraphics[width=\linewidth]{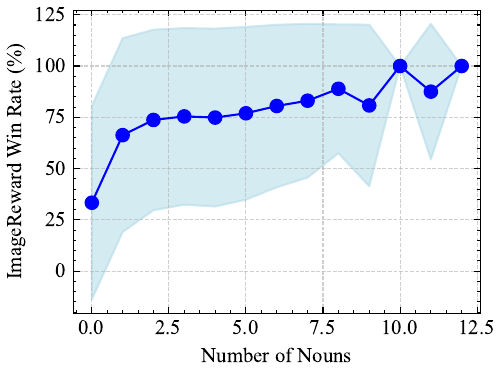}}
    \caption{Impact of the noun count in text prompts on ImageReward score~\cite{xu2023imagereward} gap and win rate between SD3 and SD2.1.}
    \label{complex-prompt}
\end{figure}

In practice, user requests involve not only complex ones but also simple ones. One intuitive factor influencing request complexity is the number of entities that need to be generated in images, i.e. the number of nouns in text prompts. As shown in \cref{complex-example}, \ref{complex-prompt}, as noun count increases, the large model is more likely to outperform the small model in quality, with quality differences becoming more noticeable. However, while large models generally excel in overall quality, small models may perform better on specific prompts. This likelihood increases as the prompt becomes simpler.
In such cases, the purely large model serving not only fails to improve performance but also incurs high cost. Therefore, it is necessary to design a routing mechanism to select the most suitable model based on the prompt's characteristics. Intuitively, the routing mechanism should direct simpler prompts to the cost-effective edge model and more complex prompts to the high-quality cloud model, thus maximizing generation quality while minimizing cost.

Formally, a T2I routing framework $R: \mathcal{X}\rightarrow \{0,1\}$ assigns user prompts $\mathcal{X}$ to the cloud large model $\mathcal{M}_c$ as 1 and to the edge light-weight model $\mathcal{M}_e$ as 0. The generated images after routing can be expressed as $\mathcal{I}_r = R(\mathcal{X})\mathcal{I}_c + (1-R(\mathcal{X}))\mathcal{I}_e$, where $\mathcal{I}_c$ and $\mathcal{I}_e$ denote the images generated using $\mathcal{M}_c$ and $\mathcal{M}_e$, respectively. Given a quality scoring function $\mathcal{Q}(\cdot)$ of the generated images, the cost budget $\tau_{fee}$ of cloud model serving and edge-cloud communication, and the response latency constraint $\tau_{time}$, the optimization objective is maximizing the quality under the budget and latency constraints, formulated as
\begin{align}
\begin{aligned}
    \max\ &R(\mathcal{X})\mathcal{Q}(\mathcal{I}_c) + (1-R(\mathcal{X}))\mathcal{Q}(\mathcal{I}_e)\\
    \text{s.t.}\ &\mathbb{P}\{R(\mathcal{X})=1\} \cdot F_c \leq \tau_{fee}\\
    &\mathbb{P}\{R(\mathcal{X})=1\} \cdot D_{\mathcal{M}_c} + \mathbb{P}\{R(\mathcal{X})=0\} \cdot D_{\mathcal{M}_e} \\
    & + D_R \leq \tau_{time},
\end{aligned}
    \label{routing_R}
\end{align}
where $D_{\mathcal{M}_c}$, $D_{\mathcal{M}_e}$, and $D_R$ denote the latency of cloud large model serving plus edge-cloud communication, the latency of the edge light-weight model serving, and the latency of the routing model execution. Considering the constraints are linear, the optimization objective can be approximated as imposing an upper bound $\rho_r$ on the routing rate to the cloud large model as
\begin{align}
\begin{aligned}
    \max\ &R(\mathcal{X})\mathcal{Q}(\mathcal{I}_c) + (1-R(\mathcal{X}))\mathcal{Q}(\mathcal{I}_e) \\
    \text{s.t.}\ &\mathbb{P}\{R(\mathcal{X})=1\} \le \rho_r.
\end{aligned}
    \label{short_opt}
\end{align}

Due to the lack of a universal standard for evaluating the quality of generated images, a multi-dimensional quality scoring function $\mathcal{Q}: \mathcal{I}\rightarrow[0,1]^N$ is required for a more comprehensive comparison. Thus, different from existing routing for typical classification and text generation tasks, the T2I routing framework's optimization objective is a new multi-objective problem. In addition, to predict quality changes in the larger image output space from the smaller text input space, we conduct a detailed analysis of how different prompt tokens influence the generated images and their multi-dimensional quality metrics, by examining the interaction between text and images during the T2I generation process. In what follow, we define the multi-metric quality measures for routing objectives in \cref{multi-di} and introduce the T2I routing framework in \cref{RouteT2I}.

\section{Multi-Metric Image Generation Quality in Routing Objectives}\label{multi-di}

The connection between text and images allows image attributes to be described through corresponding prompts. Thus, a metric can be measured by the relative relationship between an image and the positive/negative text pair describing that metric~\cite{wang2023exploring,liang2023iterative}. The contrastive quality metric for image $I$ is defined as
\begin{equation}
    q(I, m) = \sigma(CLIP(I, m^+)-CLIP(I, m^-)),
    \label{contrast_q}
\end{equation}
where $m=(m^+,m^-)$ denotes positive and negative pair, and $\sigma$ denotes the sigmoid function that transforms output values in the range from 0 to 1. According to this definition, if an image is more related to the positive prompt, and is less related to the negative prompt, the contrastive quality metric is higher. Compared to quality measure with only the positive prompt, the contrastive method evaluates whether positive or negative quality is more dominant in the image, leading to a more robust and reliable evaluation.

Considering the ambiguity and complexity of image quality, we introduce a multi-dimensional metric to comprehensively evaluate generated image quality, providing noise resistance and stability for quality-prediction-based routing. Just as real photo quality depends on factors like definition and color, our multi-metric combines traditional measures of real photos with unique aspects of generated images, such as realism and object integrity. Thus, we extend the single-metric evaluation in \cref{contrast_q} to an $N$-dimensional quality assessment of an image $I$, expressed as
\begin{equation}
    \mathcal{Q}(I)=[q(I, m_i) | i=1, 2, \ldots, N],
    \label{multi_Q}
\end{equation}
where $m_i$ is an instantiation of $m$ in \cref{contrast_q}. We adopt 10 common metrics inspired by prior works~\cite{lee2023holistic,wu2024t2i,wu2024multimodal}, with details in Appendix Sec. C. For example, for the definition metric, we use the positive prompt ``High definition photo'' and the negative prompt ``Low definition photo''; and for the object integrity, we take the positive prompt ``Object completion'' and the negative prompt ``Object twisting''.

In the multi-objective optimization setting for routing with multi-metric quality, the initial goal is to find a Pareto optimal generated image that excels in all the metrics. However, this is challenging in practice due to the potential nonexistence of such an image generated from the cloud or edge model. We thus relax the constraint of Pareto optimality by allowing suboptimal performance in some metrics if the image significantly outperforms in others when selecting higher-quality images. We define Pareto relative superiority (PRS) to quantify the quality advantage of the images $\mathcal{I}_e$ generated by the edge model over the images $\mathcal{I}_c$ generated by the cloud model. In particular, we first normalize the quality distance for the metric $i$ between an edge-generated image $I_e \in \mathcal{I}_e$ and a cloud-generated image $I_c\in\mathcal{I}_c$ as
\begin{equation}
    D_i(I_e, I_c) = \sigma \left(\frac{q(I_e,m_i) - q(I_c,m_i)}{\Gamma |\mu_i(\mathcal{I}_e) - \mu_i(\mathcal{I}_c)|}\right),
\end{equation}
where $\mu_i(\cdot)$ is the average quality of the set of cloud or edge generated images, while the sigmoid function $\sigma$ and the temperature parameter $\Gamma$ are used to modulate the data distribution, effectively distinguishing similar qualities and preventing centralization. We then define the overall quality gap as a weighted sum of distances across all $N$ metrics:
\begin{equation}
    PRS(I_e, I_c) = \sum_{i=1}^N w_iD_i(I_e, I_c),
\end{equation}
where $w_i$ denotes the weight of metric $i$, and $\sum_iw_i=1$. We can evaluate the relative quality of $I_e$ compared to $I_c$ by examining how much Pareto relative superiority deviates from 0.5. This evaluation helps to route a user prompt to the model that generates an image with higher overall quality.
\section{Design of \texttt{RouteT2I}}\label{RouteT2I}

 \begin{figure}[!t]
 \centering
 \includegraphics[width=\linewidth]{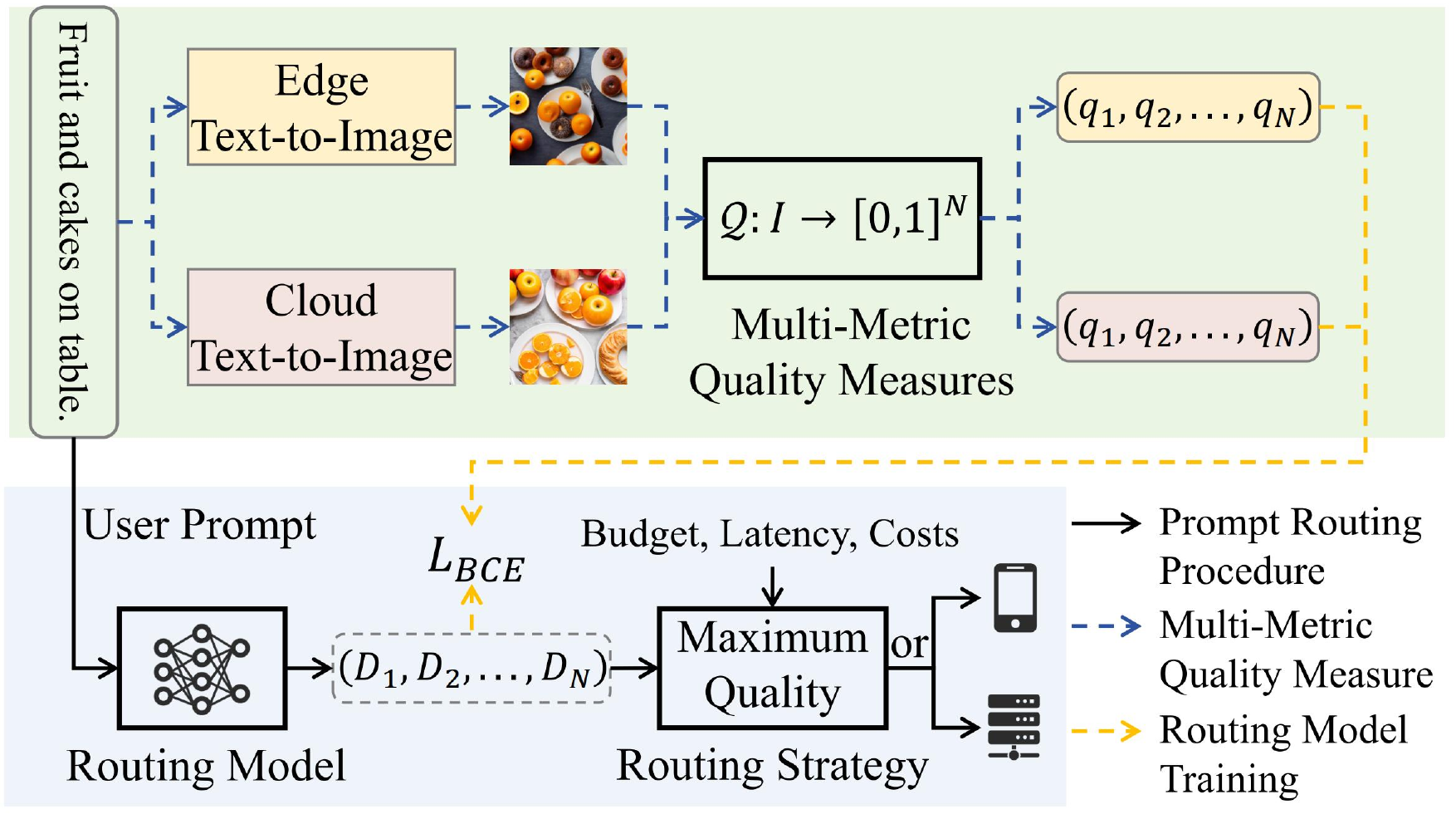}
 \caption{Overview of \texttt{RouteT2I}. \texttt{RouteT2I} assembles a pair of off-the-shelf edge and cloud text-to-image models and evaluates them using multi-metric image quality. \texttt{RouteT2I} utilizes Pareto relative superiority between qualities as supervision to train the routing model. Then, to balance quality and cost, the routing strategy determines the most suitable model for each user prompt, choosing between the edge or cloud model.}
 \label{overview}
 \end{figure}

\begin{figure*}[htbp]
    \centering
    \subcaptionbox{Token selection gate.}[0.38\linewidth]{\includegraphics[width=0.9\linewidth]{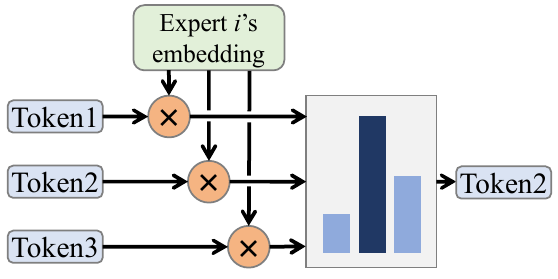}}
    \subcaptionbox{Dual-gate MoE.}[0.52\linewidth]{\includegraphics[width=0.9\linewidth]{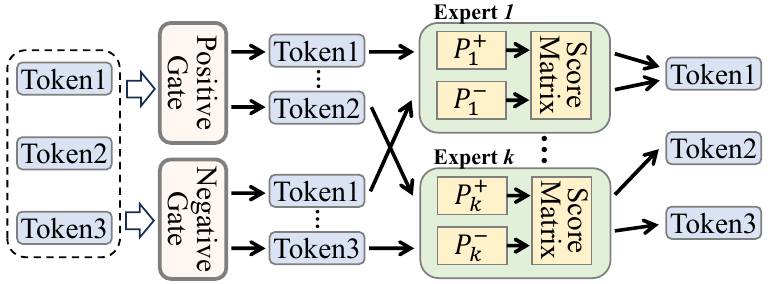}}
    \caption{Architecture of a dual-gate MoE that uses the token selection gate as the gate network. First, the gate selects the most relevant tokens for each expert. Then, experts project the tokens selected by positive and negative gates into their respective spaces and evaluate them using score matrices. Finally, the evaluations of the tokens are gathered and compared to form new tokens.}
    \label{Dual-Gate-Token-Selection-MoE}
\end{figure*}

To achieve the optimization objective in \cref{short_opt}, we propose a text-to-image model routing framework \texttt{RouteT2I}. As shown in \cref{overview}, \texttt{RouteT2I} comprises a routing model that predicts the multi-metric generation quality given user prompts and a routing strategy that selects cloud or edge model. In particular, (1) the routing model predicts the quality relationship between images generated on the edge and cloud, namely, Pareto relative superiority. Inspired by the role of prompts in generation, where prompt as sequences of tokens interact with images through cross-attention to determine the image content, we treat prompt as a set of tokens in the routing model and select key tokens with influential contextual information for each quality metric. Then, the dominant influence of these tokens is accessed by comparing their positive and negative effects; and (2) the routing strategy aims to maximize quality under cost constraints. The key idea is to use Pareto relative superiority to describe the quality disparity of images generated by the cloud and the edge models. Only prompts that show a notable quality gap when generated in the cloud are routed there, while the rest remain on the cost-effective edge.

\subsection{Routing Model Architecture Design with Dual-Gate Token Selection MoE}

To predict Pareto relative superiority between the images generated by the cloud model and the edge model from user prompts, we design a dual-gate token selection mixture-of-experts (MoE), as shown in \cref{Dual-Gate-Token-Selection-MoE}, and introduce it into Transformer network to replace linear layers.

Our model treats the user prompt as a sequence of tokens, whose influences on the image quality differ due to their weights in cross-attention during generation. To focus on key tokens for each metric, we design a token selection gate, where experts align with quality metrics and actively choose the most relevant tokens to simulate metric attention to different image attributes. Only tokens that significantly impact the quality metrics corresponding to the experts are selected, while those with minimal influences are disregarded, reducing potential interference.

Regarding the design of a dual-gate MoE, both positive and negative gates are introduced to distinguish the opposing aspects of token influences on image quality. Similar to images that have both superior and inferior regions when evaluated by metrics, these tokens are assessed by the corresponding experts for their positive and negative influences to each metric. Additionally, we introduce a comparison between two gates to identify which influence of each token predominates overall. We further improve the structure of expert to handle both positive and negative influences on a single metric simultaneously.

To support multi-metric quality prediction, our design incorporates multiple prediction heads within the model architecture. Each head outputs the prediction of a specific quality metric. Such multi-head design not only enhances noise resistance, thereby increasing robustness, but also facilitates a comprehensive evaluation of quality.

\subsubsection{Token Selection Gate}\label{token-gate}

In the token selection gate, the dot product between the token representations $T\in\mathbbm{R}^{n\times d}$ and the expert embeddings $E\in\mathbbm{R}^{k\times d}$ is computed as an affinity matrix:
\begin{align}
    A = \text{Softmax}(T\cdot E^T),
\end{align}
where $n$ denotes the number of tokens in user prompt, $k$ denotes the number of experts, $d$ denotes the hidden dimension, and the Softmax function is applied along the expert axis. $A[t, i]$ in the affinity matrix captures the correlation between token $t$ and expert $i$. Then, top-$K$ tokens with the highest relevance for each expert are selected 
\begin{align}
    M = \textrm{Top-K}(A),
\end{align}
where $M\in\mathbbm{R}^{n\times k}$ is the binary mask matrix such that $M[t,i]=1$ denotes token $t$ is selected by expert $i$, and $M[t,i]=0$ denotes not selected. Experts tend to have similar weights because the metrics they are responsible for are weighted similarly. This ensures that the tokens selected by these experts possess significant global importance. This practice minimizes interference from tokens that are insignificant across all metrics. The influences of key tokens are further evidenced by their frequent selection through the top-$K$ function and their corresponding scores within the affinity matrix. Moreover, by allocating an equal number of $K$ tokens to each expert, workloads remain balanced, and bias in quality dimensions is avoided, ultimately resulting in more comprehensive and reliable quality prediction. Overall, the token selection gate is summarized as
\begin{align}
    A, M = G(T, E),
\end{align}
where $G$ denotes the gate network.

\subsubsection{Dual-Gate MoE}

Both a positive gate $G^+$ and a negative gate $G^-$ are introduced for different experts to select tokens that have positive and negative impacts on their corresponding quality metrics, while capturing the weights of their influences. The token selection results are denoted as 
\begin{align}
    A^o, M^o = G^o(T, E^o),\ \text{where } o\in\{+,-\}.
\end{align}
The affinity matrix $A^o$ and the binary mask matrix $M^o$ are captured by the gate $G^o$ for the positive or negative aspect using the corresponding expert embedding $E^o$. The tokens selected by two gates are then evaluated by the corresponding positive or negative experts. 

To eliminate the redundancy of having separate experts for evaluating both positive and negative aspects of the same quality metric, the structure of experts is revised to allow an expert to assess both aspects simultaneously. This revision acknowledges that both positive and negative evaluations are inherently part of the same quality metric. Specifically, the original linear layer $W_i\in\mathbbm{R}^{d\times h}$ of expert on metric $i$ is split into two projection matrices $P_i^+, P_i^-\in\mathbbm{R}^{d\times l}$ and one score matrix $S_i\in\mathbbm{R}^{l\times h}$. Two projection matrices are specific to positive and negative aspects, projecting tokens into the corresponding representations in an $l$-dimensional low-dimensional space. Since $l\ll h, d$, the parameter size and the computation cost are reduced from $O(hd)$ to $O(l(h+d))$, allowing an expert to efficiently focus on both positive and negative aspects of a metric with a low cost. The shared score matrix then evaluates the impact of these representations on metric $i$.
Thus, the impact of $t$-th token $T[t]\in\mathbbm{R}^{d}$ on quality metric $i$ for positive or negative aspect $o$ is described as
\begin{align}
    &T_i^o[t] = T[t]\cdot P_i^o \cdot S_i,\ \text{where}\ o\in\{+,-\}.
\end{align}
By separating the projection matrices and score matrix, we can evaluate both positive and negative impacts on a metric within the same expert.
Then, the overall positive or negative impact of the $t$-th token on image quality can be obtained by a weighted sum of its impacts for corresponding aspects across all metrics
\begin{align}
    &T^o[t] = \sum_{i=0}^k \lambda_{t,i}^o \cdot T_i^o[t],\ \text{where}\ o\in\{+,-\}.
\end{align}
Here, $\lambda_{t,i}^o = \frac{\mathbbm{1}\{M^o[t,i]=1\}A^o[t,i]}{\sum_{j=0}^k\mathbbm{1}\{M^o[t, j]=1\}A^o[t, j]}$ denotes the normalized affinity of the $i$-th expert for the $t$-th token. By contrasting the impact of a token from both positive and negative aspects, we can assess the token's predominant influence on the image quality as
\begin{align}
    \hat{T}[t] = \sigma(T^+[t] - T^-[t]).
\end{align}
The introduction of contrasting helps diminish ambiguity in predictions, especially when tokens exhibit influences on multiple metrics from both positive and negative aspects.

\subsection{Routing Strategy Design}

Given the Pareto relative superiority predicted by routing model, we propose a routing strategy to effectively route user prompts to suitable models to achieve the trade-off between quality and costs. With a preset routing rate, we can determine the proportion of user prompts sent to the cloud, focusing on those where the cloud model significantly outperforms the edge model in quality. Since Pareto relative superiority represents this quality gap, we can efficiently filter prompts by setting a threshold $\alpha$ on it. In particular, the prompts with a Pareto relative superiority above the threshold are better handled by the edge model for cost efficiency, while those below the threshold achieve superior quality when processed in the cloud. Thus, the optimization objective in \cref{short_opt} can be expressed as:
\begin{align}
    \max\limits_{\alpha\le1/2} \mathbbm{P}\{PRS(I_e, I_c) < \alpha\ |\ I_e, I_c \in \mathcal{I}_e, \mathcal{I}_c \} \le \rho_r.
\end{align}
Since Pareto relative superiority being greater than or less than 1/2 indicates a relative advantage or disadvantage, respectively, we set a upper bound 1/2 on $\alpha$ to prevent user prompts with images generated at the edge of better quality from being routed to the cloud.

\section{Experiment}

\subsection{Experimental Setup}

\begin{table*}[]
\centering
\renewcommand{\arraystretch}{0.8}
\resizebox{\linewidth}{!}{
\begin{tabular}{l|cccccccccc|c}
\toprule
\multirow{2}{*}{Router} & \multicolumn{10}{c|}{Image Quality Metrics}                                                             & \multirow{2}{*}{$\Delta P(\%)$} \\
                        & Definition & Detail & Clarity & Sharpness & Harmony & Realism & Color  & Consistency & Layout & Integrity &                   \\  \midrule
Edge Model              & 0.6251          & 0.6685          & 0.6076          & 0.6537          & 0.5949          & 0.5575          & 0.4680          & 0.5088          & 0.4860          & 0.4690          & -           \\
Cloud Model             & 0.6337          & 0.6847          & 0.6346          & 0.6703          & 0.5930          & 0.5868          & 0.5134          & 0.5199          & 0.5345          & 0.4972          & -          \\ \hdashline
Random                  & 0.6294          & 0.6766          & 0.6211          & 0.6620          & 0.5939          & 0.5721          & 0.4907          & 0.5144          & 0.5102          & 0.4831          & 40.00          \\
RouteLLM-BERT~\cite{ong2024routellm} & 0.6347          & 0.6792          & 0.6305          & 0.6651          & 0.5960          & 0.5788          & 0.4982          & 0.5160          & 0.5167          & \textbf{0.4866} & 71.51          \\
RouteLLM-MF~\cite{ong2024routellm} & \textbf{0.6364} & \textbf{0.6814} & 0.6299          & 0.6660          & 0.5952          & 0.5776          & 0.4970          & 0.5164          & 0.5149          & 0.4850          & 69.90          \\
Hybrid LLM~\cite{ding2024hybrid}            & 0.6327          & 0.6784          & 0.6306          & 0.6677          & 0.5964          & 0.5787          & 0.5008          & 0.5161          & \textbf{0.5191} & 0.4864          & 73.49          \\
ZOOTER~\cite{lu2023routing} & 0.6350          & 0.6796          & 0.6315          & 0.6672          & 0.5966          & 0.5788          & 0.5004          & 0.5166          & 0.5179          & 0.4854          & 77.95          \\
\rowcolor[gray]{0.9}
\texttt{RouteT2I} (Ours)   & 0.6350          & 0.6786          & \textbf{0.6318} & \textbf{0.6679} & \textbf{0.5975} & \textbf{0.5804} & \textbf{0.5010} & \textbf{0.5167} & 0.5189          & 0.4865          & \textbf{83.97} \\ \bottomrule
\end{tabular}   }
\caption{The multi-dimensional quality of images generated by text-to-image models, with the router selecting between edge and cloud models for each prompt at a 50\% routing rate. The higher the metrics, the better.}
\label{main_results_multi-metrics}
\end{table*}

{\bf Dataset.} We take COCO2014~\cite{lin2014microsoft}, a comprehensive dataset for object detection, segmentation, and captioning tasks. We select its captions as user prompts $\mathcal{X}$. Given these prompts, we generate images $\mathcal{I}_e, \mathcal{I}_c$ using different open-source text-to-image models. Similarly, we use the LAION dataset~\cite{schuhmann2022laion}, with results shown in Appendix Sec. F.

{\bf Models.} We take T2I models with varying performances and sizes for cloud and edge usage, include Stable Diffusion 1.5 (SD1.5)~\cite{rombach2022high}, Stable Diffusion 2.1 (SD2.1)~\cite{rombach2022high}, Stable Diffusion XL (SDXL)~\cite{podell2023sdxl}, Stable Diffusion XL-Refiner (XL-Refiner)~\cite{podell2023sdxl}, Stable Diffusion 3 (SD3)~\cite{esser2024scaling}, PixArt-$\alpha$~\cite{chen2023pixart}, and Infinity~\cite{han2024infinity}. 
These models use various text encoders, such as CLIP~\cite{radford2021learning}, OpenCLIP~\cite{cherti2023reproducible}, and T5~\cite{raffel2020exploring}, and are based on architectures like diffusion, diffusion Transformer and autoregressive. Details are in Appendix Sec. A.


{\bf Baselines and Settings.} We first introduce {\em random routing} as a baseline, where prompts are randomly assigned to candidate models, given a specified routing rate to the cloud. Additionally, although there does not exist any previous work on text-to-image generation routing, for a comprehensive comparison, we also reproduce several representative routing methods for LLM by adapting them to our scenarios, including RouteLLM~\cite{ong2024routellm} with BERT classifier or matrix factorization, Hybrid LLM~\cite{ding2024hybrid}, and ZOOTER~\cite{lu2023routing}.
Specifically, we keep their text input processing and quality threshold design, but replace their optimization goal with the average image quality across metrics as they only handle single-objective optimization.

Unless specified in their paper, the routing models are based on Transformers and take the same hyperparameters as in our method. All routers are implemented in PyTorch 2.3.1 and trained using Adam optimizer with a learning rate of 2e-5, a weight decay of $0$, a batch size of $16$, on a NVIDIA 4090D for about 10 epochs.

{\bf Metrics.} For cost efficiency, we introduce \textit{routing rate}, namely, the proportion of user prompts to the cloud model:
\begin{align}
    p = \mathbbm{P}\{R(\mathcal{X})=1\}.
\end{align}
Smaller routing rate indicates better cost efficiency.

For image generation quality, the absolute performance of routing is constrained by the gap between edge and cloud T2I models. Thus, we introduce a set of relative performance metrics for evaluation. We first fix a routing rate $p$ and introduce the \textit{win rate} of the selected models over the cloud model after routing:
\begin{align}
    w(\mathcal{I}_{r,p}) &= \mathbbm{P}\{ \mathcal{Q}(I_r) \ge \mathcal{Q}(I_c)\ |\ I_r,I_c\in\mathcal{I}_{r,p}, \mathcal{I}_c \},
\end{align}
where $\mathcal{I}_{r,p}$ denotes the generated images after routing at a routing rate $p$. 
To intuitively show the improvement in win rate, we introduce the \textit{normalized win rate improvement}, which is the increase in win rate achieved by a routing method compared to the baseline, divided by the increase achieved by the optimal oracle routing. A higher value indicates that the routing method is closer to the oracle routing:
\begin{align}
    \Delta \overline{w}(p) = \frac{w(\mathcal{I}_{r,p}) - w(\mathcal{I}_{b,p})}{w(\mathcal{I}_{o,p}) - w(\mathcal{I}_{b,p})},
\end{align}
where $\mathcal{I}_{b,p}$ and $\mathcal{I}_{o,p}$ denote the generated images using the random baseline and oracle, respectively. 
To quantify the improvement of image quality metrics, we introduce {\em relative performance improvement}, which measures the proportion of the performance gain achieved through routing compared to the edge model, relative to the total possible improvement brought by fully using the cloud model:
\begin{align}
    \Delta P = \frac{1}{N} \sum_{i=1}^N \frac{\mu_i(\mathcal{I}_r) - \mu_i(\mathcal{I}_e)}{|\mu_i(\mathcal{I}_c) - \mu_i(\mathcal{I}_e)|}.
\end{align}
This metric effectively quantifies routing effectiveness while accounting for the original quality gap.

For cost-quality balance, we introduce the \textit{cost saving} to measure the reduction of the routing rate to the cloud model, compared to the random baseline at a given relative performance improvement $\Delta P$:
\begin{align}
    \gamma(\Delta P) = \frac{p_{b}(\Delta P) - p_{r}(\Delta P)}{p_{b}(\Delta P)},
\end{align}
where $p_b(\Delta P)$ and $p_r(\Delta P)$ represent the routing rates of baseline and router at a given $\Delta P$, respectively.

\subsection{Main Results}

We present the routing performance of our \texttt{RouteT2I} and baselines using SD3 as the cloud model and SD2.1 as the edge model. Human evaluation, cost analysis, visual results, and more T2I pair results are provided in Appendix.


    \begin{table}[]
        \centering
        \renewcommand{\arraystretch}{0.8}
        \resizebox{0.9\linewidth}{!}{
        \begin{tabular}{l|ccccc}
            \toprule
            \diagbox{Router}{$p$} & 40\%           & 50\%           & 60\%           & 70\%           & 80\%           \\       \midrule
            RouteLLM-BERT~\cite{ong2024routellm} & 24.29          & 22.45          & 19.07          & 17.62          & 20.59          \\
            RouteLLM-MF~\cite{ong2024routellm} & 25.65          & 23.29          & 19.92          & 16.67          & 20.59          \\
            Hybrid LLM~\cite{ding2024hybrid} & 23.77          & 19.75          & 14.92          & 13.62          & 16.09          \\
            ZOOTER~\cite{lu2023routing} & 26.77          & 21.97          & 17.97          & 16.97          & 21.04          \\
            \rowcolor[gray]{0.9}
            \texttt{RouteT2I} (Ours)         & \textbf{30.60} & \textbf{25.81} & \textbf{20.32} & \textbf{18.02} & \textbf{21.94} \\ \bottomrule
        \end{tabular}
        }
        \caption{Normalized win rate improvements $\Delta \overline{w}$ (\%) at different routing rates, i.e. the ratio of improvement to oracle improvement.}
        \label{main_result_wining-rate}
    \end{table}

\textbf{Win Rates.} \cref{main_result_wining-rate} shows our \texttt{RouteT2I} demonstrates a significant improvements on normalized win rate improvements across all routing rates. Specifically, at a 40\% routing rate, our improvement reaches 30.60\%, exceeding the baselines by at least 3.83\% and demonstrating strong consistency with the oracle.


\textbf{Image Quality.} \cref{main_results_multi-metrics} shows \texttt{RouteT2I} outperforms all baselines in 6 of 10 metrics. Our multi-metric optimization balances all metrics simultaneously, while others focus on just a few, often at the cost of others. We also achieve the highest relative performance improvement, reaching 83.97\% of cloud model's gain over edge model, surpassing others by at least 6\%. Experiments in Appendix Sec. H show our method may lead to performance gains beyond those of the cloud model itself and is effective for T2I models with various text encoders and architectures.

\begin{table}[]
\centering
\renewcommand{\arraystretch}{0.8}
\resizebox{0.9\linewidth}{!}{
\begin{tabular}{l|ccccc}
\toprule
\diagbox{Router}{$\Delta P$} & 40\%           & 50\%           & 60\%           & 70\%           & 80\%           \\       \midrule
RouteLLM-BERT~\cite{ong2024routellm} & 56.15 & 51.39 & 46.92 & 42.70 & 40.21 \\
RouteLLM-MF~\cite{ong2024routellm} & 48.86 & 49.90 & 48.20 & 44.89 & 41.50 \\
Hybrid LLM~\cite{ding2024hybrid} & 62.06 & 58.85 & 53.63 & 49.92 & 33.38 \\
ZOOTER~\cite{lu2023routing}  & 69.28 & 65.76 & 60.81 & 57.35 & 49.64 \\
\rowcolor[gray]{0.9}
\texttt{RouteT2I} (Ours)          & \textbf{71.81} & \textbf{70.24} & \textbf{66.61} & \textbf{60.01} & \textbf{53.53} \\ \bottomrule
\end{tabular}
}
\caption{Cost saving $\gamma$ (\%) under a given relative performance improvements ($\Delta P$) target. $\gamma$ is the ratio of savings to the routing rate required by the random baseline.}
\label{cost-saving}
\end{table}

\textbf{Cost Efficiency.} \cref{cost-saving} presents the cost savings at given relative performance improvement targets. At a 40\% relative performance improvement, \texttt{RouteT2I} reduce cloud service needs by up to 71.81\% of the calls required by random routing and by at least 5.80\% compared to other baselines. In other cases, \texttt{RouteT2I} still reduces over 50\% costs, demonstrating high cost efficiency.

\begin{figure}
    \centering
    \subcaptionbox{}[0.42\linewidth]{\includegraphics[width=\linewidth]{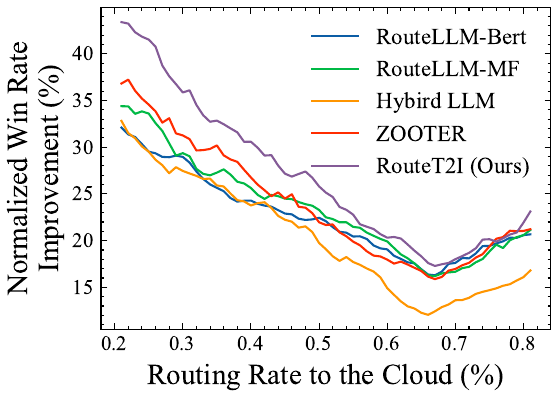}}
    \subcaptionbox{\label{RPI}}[0.42\linewidth]{\includegraphics[width=\linewidth]{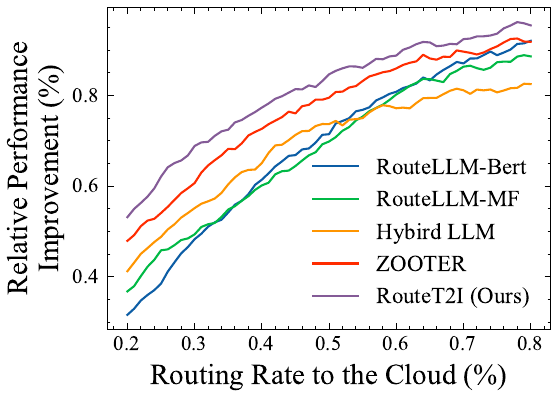}}
    \caption{Visualization of (a) Normalized win rate improvement $\Delta \overline{w}$ and (b) relative performance improvement $\Delta P$ at varying routing rates $p$ to the cloud.}
    \label{visualization_winrate}
\end{figure}

\textbf{Visualization.} As shown in \cref{visualization_winrate}, \texttt{RouteT2I} performs well across routing rates, demonstrating its ability to predict quality gaps between T2I models and route effectively.

Visual examples are shown in Appendix Sec. G.


\subsection{Routing Performance on More T2I Model Pairs}

\begin{table}[]
\centering
\renewcommand{\arraystretch}{0.8}
\resizebox{\linewidth}{!}{
\begin{tabular}{ll|rrrrr}
\toprule
\multicolumn{2}{c|}{\diagbox{Cloud-Edge}{$p$}}   & 40\%  & 50\%  & 60\%  & 70\%  & 80\%  \\   \midrule
SD3                          & XL-Refiner                  & 27.42 & 22.39 & 18.05 & 16.85 & 20.89 \\
SD3                          & SDXL                        & 29.78 & 26.81 & 22.35 & 16.98 & 21.30 \\
SD3                          & SD1.5                       & 26.24 & 20.88 & 16.19 & 14.04 & 19.22 \\
SD2.1                        & XL-Refiner                  & 21.62 & 18.31 & 20.46 & 25.30 & 28.66 \\
SD2.1                        & SDXL                        & 19.46 & 16.87 & 17.27 & 20.42 & 23.69 \\
SD2.1                        & SD1.5                       & 10.77 & 9.28  & 10.12 & 12.22 & 15.76 \\
XL-Refiner                   & SDXL                        & 13.57 & 10.15 & 9.74  & 11.81 & 12.59 \\
XL-Refiner                   & SD1.5                       & 19.65 & 16.98 & 18.87 & 22.79 & 25.57 \\
SDXL                         & SD1.5                       & 18.38 & 17.01 & 21.57 & 24.75 & 32.49 \\ 
SD3                          & PixArt-$\alpha$             & 32.26 & 27.59 & 24.74 & 30.73 & 38.97 \\
PixArt-$\alpha$              & SD2.1                       & 25.82 & 21.26 & 22.55 & 28.19 & 33.76 \\
PixArt-$\alpha$              & SDXL                        & 16.71 & 13.57 & 12.37 & 16.35 & 22.14 \\
PixArt-$\alpha$              & XL-Refiner                  & 22.51 & 19.14 & 18.18 & 23.72 & 28.44 \\
PixArt-$\alpha$              & SD1.5                       & 22.38 & 20.52 & 19.93 & 25.40 & 32.43 \\
SD3                          & Infinity                    & 22.19 & 19.97 & 17.77 & 19.81 & 25.39 \\
SD2.1                        & Infinity                    & 21.98 & 19.86 & 22.60 & 28.45 & 36.18 \\
PixArt-$\alpha$              & Infinity                    & 27.69 & 23.48 & 24.12 & 30.89 & 38.91 \\ \bottomrule
\end{tabular}
}
\caption{Normalized win rate improvement $\Delta \overline{w}$ (\%) over oracle routing of our \texttt{RouteT2I} with different cloud-edge model pairs.}
\label{Performance_diff_device-cloud-model}
\end{table}


To test our method across T2I model pairs, \cref{Performance_diff_device-cloud-model} shows \texttt{RouteT2I}'s performance with various cloud and edge models. Results reveal significant improvements, especially for pairs with large quality gaps, such as SD3 on cloud and others on edge, where \texttt{RouteT2I} achieves over 25\% of the oracle routing's improvement at a 40\% routing rate. Even with similar model pairs like XL-Refiner and SDXL, it reaches 10\% of the oracle's improvement, demonstrating its ability to discern subtle differences between T2I models.

\subsection{Ablation Experiment}

\begin{table}[]
\centering
\renewcommand{\arraystretch}{0.75}
\resizebox{0.9\linewidth}{!}{
\begin{tabular}{l|ccccc}
\toprule
\diagbox{Router}{$p$}   & 40\%  & 50\%  & 60\%  & 70\%  & 80\%  \\   \midrule
w/o Multi-Metric        & 27.37 & 22.81 & 18.87 & 16.47 & 19.92 \\
w/o Token Selection     & 27.82 & 23.05 & 18.27 & 16.77 & 19.24 \\
w/o Dual-Gate           & 27.22 & 22.09 & 19.27 & 17.37 & 21.62 \\
\texttt{RouteT2I}                & \textbf{30.60} & \textbf{25.81} & \textbf{20.32} & \textbf{18.02} & \textbf{21.94} \\ \bottomrule
\end{tabular}
}
\caption{Ablation experiment of \texttt{RouteT2I}, showing normalized win rate improvement $\Delta \overline{w}$ (\%) on given routing rates.}
\label{Ablation}
\end{table}

\cref{Ablation} presents ablation results for \texttt{RouteT2I}'s multi-metric quality optimization, token selection gate, and dual-gate MoE. The multi-metric quality optimization allows routers to comprehensively assess the image quality gap, significantly enhancing overall performance. Removing this results in a performance decline across all routing rates, particularly at 40\% routing rate, where the drop is approximately 3.23\%. The token selection gate prioritizes critical tokens, ensuring routing focuses on key quality factors while reducing interference from irrelevant elements. Its removal notably impacts high routing rates, causing a performance decrease of around 2\% at 80\% routing rate. The dual-gate MoE effectively distinguishes subtle differences between cloud and edge model outputs, proving most effective at medium routing rates, especially near 50\%.

\section{Conclusion}


In this work, we have proposed a new text-to-image model routing design \texttt{RouteT2I} between edge and cloud. \texttt{RouteT2I} adopts the routing model with a dual-gate token selection MoE to predict how user prompts affect multiple image quality metrics by identifying and evaluating key tokens with contrastive method. \texttt{RouteT2I} further introduces a routing strategy that exploits the predicted advantage of the edge model over the cloud model to determine which side for image generation. Extensive evaluation has demonstrated that \texttt{RouteT2I} can significantly enhance generation quality at a specified routing rate and meanwhile reduce cost at a given quality target.

\section*{Acknowledge}
This work was supported in part by National Key R\&D Program of China (No. 2022ZD0119100), China NSF grant No. 62202297, No. 62025204, No. 62202296, No. 62272293, No. 62441236, and No. U24A20326, SJTU-Huawei Research Program, and Tencent WeChat Research Program. The opinions, findings, conclusions, and recommendations expressed in this paper are those of the authors and do not necessarily reflect the views of the funding agencies or the government.

{
    \small
    \bibliographystyle{ieeenat_fullname}
    \bibliography{main}
}

\clearpage
\appendix
\setcounter{page}{1}
\setcounter{equation}{0}
\setcounter{figure}{0}
\setcounter{table}{0}
\maketitlesupplementary

\begin{table}[]
\centering
\resizebox{\linewidth}{!}{
\begin{tabular}{l|rrr}
\toprule
Text Encoder   & \#Parameter & Max Length (tokens) & Vocabulary Size \\  \midrule
CLIP-ViT/L     & 123.65 M    & 77                  & 49408           \\
OpenCLIP-ViT/H & 354.0 M     & 77                  & 49408           \\
OpenCLIP-ViT/G & 694.7 M     & 77                  & 49408           \\
Flan-T5-XL     & 3 B         & 512                 & 32128           \\
T5-XXL         & 11 B        & 512                 & 32128           \\ \bottomrule 
\end{tabular}
}
\caption{Text encoders of text-to-image models.}
\label{T2I_text_encoders}
\end{table}

\begin{table*}[t]
\centering
\resizebox{0.8\linewidth}{!}{
\begin{tabular}{l|llr}
\toprule
Text-to-Image Model         & Text Encoder                          & Type                  & \#Param     \\  \midrule
Stable Diffusion 1.5        & CLIP-ViT/L                            & Diffusion             & 0.86 B      \\
Stable Diffusion 2.1        & OpenCLIP-ViT/H                        & Diffusion             & 0.86 B      \\
Stable Diffusion XL         & OpenCLIP-ViT/G and CLIP-ViT/L         & Diffusion             & 2.6 B       \\
Stable Diffusion XL-Refiner & OpenCLIP-ViT/G and CLIP-ViT/L         & Diffusion             & -           \\
Stable Diffusion 3          & OpenCLIP-ViT/G, CLIP-ViT/L and T5-XXL & Diffusion Transformer & 8 B         \\
PixArt-$\alpha$             & T5-XXL                                & Diffusion Transformer & 0.6 B       \\
Infinity                    & Flan-T5-XL                            & AutoRegressive        & 2 B         \\ \bottomrule
\end{tabular}
}
\caption{Details of text-to-image models used in routing.}
\label{T2Imodels_used}
\end{table*}

\section{Details of Text-to-Image Model Choices on Edge and Cloud}\label{T2I-detailed-information}

To evaluate the performance of routing methods, we choose a series of text-to-image (T2I) models with varying sizes, computational costs, and performances on edge and cloud. This section provides a brief overview of the T2I models used in our experiments. As shown in \cref{T2Imodels_used}, we utilize diffusion-based models such as Stable Diffusion 1.5 (SD1.5), Stable Diffusion 2.1 (SD2.1), Stable Diffusion XL (SDXL), and Stable Diffusion XL-Refiner (XL-Refiner), alongside diffusion Transformer models Stable Diffusion 3 (SD3) and PixArt-$\alpha$, and an autoregressive model Infinity. Notably, Stable Diffusion XL-Refiner is a refinement model applied after image generation by SDXL to enhance image quality. We test the performance of our routing method under different edge-cloud model pairs to verify its effectiveness.

These T2I models also differ in their use of text encoders, which play a critical role in conditioning the models on textual input. As detailed in \cref{T2Imodels_used}, the models vary in both the type and number of text encoders employed. For instance, SD1.5 and SD2.1 rely on lightweight encoders such as CLIP or OpenCLIP, while SD3 incorporates three text encoders: CLIP, OpenCLIP, and T5. As shown in \cref{T2I_text_encoders}, these text encoders vary significantly in parameters and maximum token length limits. It is worth noting that the T5 encoder, used in SD3, PixArt-$\alpha$, and Infinity, is a large-scale text-to-text transfer transformer model whose size may even exceed that of the T2I model itself. This means that when comparing the relative size of T2I models, the size of the text encoder cannot be ignored and needs to be taken into account.



\section{Comparison of Input and Output Spaces for Text-to-Image Models}\label{text-image-space}

To begin, we define the sizes of text and image spaces. For simplicity, we measure the size of a space by the number of possible samples it can contain.

For the text space, models typically have a fixed vocabulary size and a maximum input length. Assuming a vocabulary size of $|\mathcal{V}|$ and a maximum input token length of $L$, the size of the text space can be approximated as:
\begin{equation}
    S_{\text{text}} = |\mathcal{V}|^L
\end{equation}

The size of the image space is determined by the image resolution and the color depth per pixel. Each pixel’s color depth defines the number of possible states for that pixel. For an RGB image with a resolution of $W\times H$ and a color depth of 24 bits (8 bits each for red, green, and blue), the size of the image space can be approximated as:
\begin{equation}
    S_{\text{image}} = (2^{\text{color\_depth}})^{W\times H}
\end{equation}

To facilitate comparison between large numerical values of space sizes, we define the scale ratio $R$ as the logarithmic of the ratio between output space size $S_{\text{out}}$ and input space size $S_{\text{in}}$:
\begin{equation}
    R=\log(S_{\text{out}}/S_{\text{in}})
\end{equation}
If $R>0$, the output space is larger than the input space; if $R<0$, the input space is smaller than the output space; and if $R\approx 0$, the two spaces are of comparable size.

For large language models (LLMs), both the input and output spaces are text-based, with similar vocabulary sizes and token limits. Therefore, the scale ratio of output-to-input space sizes can be approximated as:
\begin{equation}
    R_{\text{LLM}} = \log(S_{\text{text\_out}}/S_{\text{text\_in}}) \approx 0
\end{equation}
This indicates that the input and output spaces are roughly equal in scale.

For text-to-image models, the input space is text-based, while the output space is image-based. Assuming the text encoder uses CLIP with a vocabulary size of $|\mathcal{V}|=49408$ and a maximum input length of $L=77$ as shown in \cref{T2I_text_encoders}, and the output is a $512\times512$ RGB image with a color depth of 24 bits, the scale ratio of output-to-input space sizes can be approximated as:
\begin{align}
    R_{\text{T2I}} 
    &= \log(S_{\text{image}}/S_{\text{in}}) \\
    &= HW\cdot color\_depth \cdot\log2 - L\cdot \log(|\mathcal{V}|) \\
    &= 512\times512\times24\log2 - 77\times\log49408 \\
    &\approx 4360072
\end{align}
Similarly, for T5 text encoder with a vocabulary size of $|\mathcal{V}|=32128$ and a maximum input length of $L=512$, the scale ratio is:
\begin{align}
    R_{\text{T2I}} 
    &= \log(S_{\text{image}}/S_{\text{in}}) \\
    &= 512\times512\times24\log2 - 512\times\log32128 \\
    &\approx 4355591
\end{align}
This indicates that, for T2I models, the output image space is at least $e^{4\times10^6}$ times larger than the input text space. Considering that T2I models can generate higher-resolution images, such as $768\times768$ or $1024\times1024$, this ratio becomes even larger.

Thus, for LLMs, the input and output spaces are roughly comparable in size. In contrast, for T2I models, the output image space is significantly larger than the input text space. This means that predictive routing for T2I models must infer quality changes in a vastly larger and more complex output image space based on a constrained input text space. This significant disparity poses a greater challenge for predictive routing of T2I models. To address this, we propose a routing optimization strategy based on multi-dimensional image quality metrics to reduce noise and inaccuracies in predictions, while designing the routing model to capture the complex mapping between input and output spaces.

\section{Image Quality Metrics}\label{multi-metric-detail}

\begin{table}[]
\centering
\resizebox{\linewidth}{!}{
\begin{tabular}{l|l}
\toprule
Metrics     & Positive/Negative Text Pairs                      \\ \midrule
Definition  & (``High definition photo", ``Low definition photo") \\
Detail      & (``Detailed photo", ``Lacking Detail photo")        \\
Clarity     & (``Clear photo", ``Blurred photo")                  \\
Sharpness   & (``Sharp", ``Hazy")                                 \\
Harmony     & (``Visually harmonious", ``Visually chaotic")       \\
Realism     & (``Realism", ``Distortion")                         \\
Color       & (``Color accurate", ``Color distorted")             \\
Consistency & (``Color consistency", ``Color conflict")           \\
Layout      & (``Reasonable composition", ``Chaotic composition") \\
Integrity   & (``Object completion", ``Object twisting")         \\ \bottomrule 
\end{tabular}
}
\caption{Multi-metric image generation quality and their positive/negative text description pairs.}
\label{detail_multi_metric}
\end{table}


To comprehensively evaluate the quality of generated images, we introduce a set of metrics that include both general criteria applicable to real photos and unique indicators specific to generated images. As shown in \cref{detail_multi_metric}, our evaluation, which targets a realistic object generation task, adopts 10 commonly used metrics suggested by HEIM~\cite{lee2023holistic}, T2I-Scorer~\cite{wu2024t2i}, and VisionPrefer~\cite{wu2024multimodal}. Metrics such as clarity and sharpness are considerations for real photos but also apply to generated images, and metrics like object integrity and realism are unique to evaluating generated images. Each of these metrics is accompanied by a pair of positive and negative descriptive texts that characterize the nature of the attribute being evaluated. This multi-dimensional quality metric allows us to more comprehensively consider subtle differences between images generated by different T2I models, enabling more accurate routing decision. Furthermore, our approach is not limited to the aforementioned metrics. \cref{detail_multi_metric} merely illustrates an example. Considering that different generation scenarios have varying quality requirements, the multi-metric quality formulation and the weights for different metrics in Eq. (3) and Eq. (4) of the main text also allow tailoring to specific application needs. Image quality in our routing objective can encompass any number of dimensions and utilize arbitrary evaluation methods without requiring modifications to our routing framework.

\section{Human Preference}\label{A:human}

\begin{table}[]
\centering
\resizebox{\linewidth}{!}{
\begin{tabular}{l|cc|cc}
\toprule
Router                 & HPSv2~\cite{wu2023human}           & $\Delta P(\%)$   & MPS~\cite{zhang2024learning}             & $\Delta P(\%)$   \\  \midrule
Edge                   & 0.2725          & -                & 0.3193          & -                \\
Cloud                  & 0.2925          & -                & 0.6806          & -                \\ \hdashline
Random                 & 0.2825          & 50.00            & 0.5000          & 50.00            \\
RouteLLM-BERT~\cite{ong2024routellm}     & 0.2843          & 59.00            & 0.5058          & 51.62            \\
RouteLLM-MF~\cite{ong2024routellm}       & 0.2852          & 63.50            & 0.5138          & 53.83            \\
HybridLLM~\cite{ding2024hybrid}          & 0.2842          & 58.50            & 0.5058          & 51.62            \\
ZOOTER~\cite{lu2023routing}            & 0.2846          & 60.50            & 0.5137          & 53.81            \\
\rowcolor[gray]{0.9}
\texttt{RouteT2I} (Ours)        & \textbf{0.2864} & \textbf{69.50}   & \textbf{0.5273} & \textbf{57.57}   \\ \bottomrule
\end{tabular}   }
\caption{The alignment of images generated by text-to-image models with human judgments, with the router selecting between edge and cloud models for each prompt at a 50\% routing rate.}
\label{A:human-judge}
\end{table}

\begin{table*}[]
\centering
\resizebox{\linewidth}{!}{
\begin{tabular}{l|cccccccccc|c}
\toprule
\multirow{2}{*}{Router}     & \multicolumn{10}{c|}{Image Quality Metrics}                                                             & \multirow{2}{*}{$\Delta P(\%)$} \\
                            & Definition & Detail & Clarity & Sharpness & Harmony & Realism & Color  & Consistency & Layout & Integrity &                   \\  \midrule
Edge Model                  & 0.6922          & 0.6496          & 0.6019          & 0.6477          & 0.6284          & 0.5974          & 0.4922          & 0.5308          & 0.4835          & 0.4611          & -              \\ 
Cloud Model                 & 0.7057          & 0.6716          & 0.6326          & 0.6657          & 0.6323          & 0.6454          & 0.5334          & 0.5481          & 0.5315          & 0.4791          & -              \\ \hdashline
Random                      & 0.6990          & 0.6606          & 0.6173          & 0.6567          & 0.6304          & 0.6214          & 0.5128          & 0.5395          & 0.5075          & 0.4701          & 50.00          \\
RouteLLM-BERT~\cite{ong2024routellm}          & \textbf{0.7073} & 0.6674          & 0.6226          & 0.6591          & 0.6331          & 0.6257          & 0.5150          & 0.5424          & 0.5130          & 0.4707          & 73.89          \\
RouteLLM-MF~\cite{ong2024routellm}            & 0.7067          & \textbf{0.6680} & 0.6240          & 0.6619          & 0.6326          & 0.6276          & 0.5159          & 0.5422          & 0.5131          & 0.4709          & 75.21          \\
Hybrid LLM~\cite{ding2024hybrid}              & \textbf{0.7073} & 0.6670          & 0.6235          & 0.6613          & 0.6331          & 0.6273          & 0.5169          & 0.5428          & 0.5138          & 0.4712          & 76.80          \\
ZOOTER~\cite{lu2023routing}                 & 0.7053          & 0.6657          & 0.6222          & 0.6619          & 0.6337          & 0.6285          & 0.5167          & 0.5427          & 0.5131          & \textbf{0.4719} & 76.45          \\
\rowcolor[gray]{0.9}
\texttt{RouteT2I} (Ours)    & 0.7063          & 0.6677          & \textbf{0.6241} & \textbf{0.6624} & \textbf{0.6344} & \textbf{0.6291} & \textbf{0.5175} & \textbf{0.5437} & \textbf{0.5145} & 0.4710          & \textbf{81.38} \\ \bottomrule
\end{tabular}   }
\caption{The multi-dimensional quality of images generated by text-to-image models on the LAION dataset, with the router selecting between edge and cloud models for each prompt at a 50\% routing rate. The higher the metrics, the better.}
\label{A:laion-main}
\end{table*}

We evaluate alignment with human judgments using the SOTA Human Preference Score (HPSv2)~\cite{wu2023human} and Multi-dimensional Human Preference (MPS)~\cite{zhang2024learning}, whose scoring models were trained on human preference datasets. Although our routing is not specifically optimized for these metrics, it still achieves 19.50\% and 7.57\% improvements over random routing, as shown in \cref{A:human-judge}, on $\Delta P$, which quantifies how much of the quality gap between the edge and cloud models is recovered by routing. These results show that our routing generalizes well and aligns closely with human preferences.

\section{Training Cost and Inference Latency}\label{A:infer_cost}


Our routing model, comprising 58.17 million parameters, can be trained in 7 minutes on a single NVIDIA RTX 4090D GPU. During training, it efficiently utilizes only 2.7GB of system RAM and 2.5GB of GPU VRAM, demonstrating strong computational efficiency.
For edge deployment, the model shows promising performance on embedded platforms. On the NVIDIA Jetson TX2, inference takes an average of 64.5ms per image, using 4.0GB RAM and 1.6GB swap space. On the less powerful NVIDIA Jetson Nano, the inference time increases to 131.3ms per image, using 1.1GB RAM and 2.7GB swap. These results highlight the efficiency of our routing model and its suitability for resource-constrained edge devices, showing that on-device routing inference incurs only a small overhead while achieving notable performance improvements.


\section{Results on Other Datasets}\label{A:dataset}

We also conduct experiments on a subset of the public LAION2B-en-aesthetic dataset~\cite{schuhmann2022laion}, consisting of 20k samples for training and 10k for validation. Prompts from both the COCO and the LAION datasets are collected from authentic human inputs. Compared to COCO prompts' concise and objective styles, LAION's are more colloquial and stylistic. \cref{A:laion-main} shows our method still performs well on the LAION dataset, demonstrating its generalization.

\section{Visual Results}\label{A:vis}

\begin{figure*}[!t]
\centering
\includegraphics[width=\linewidth]{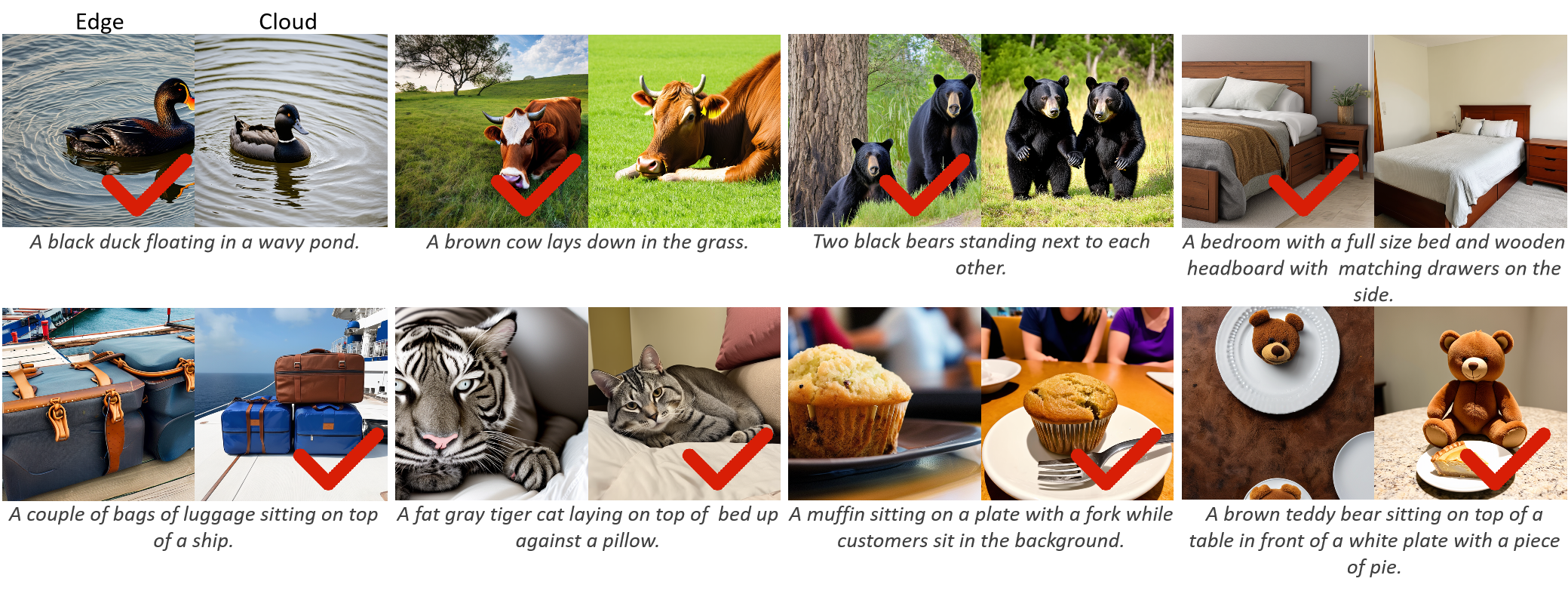}
\caption{Visual results of our \texttt{RouteT2I} at a routing rate of 50\%, using SD2.1 as the edge model and SD3 as the cloud model. The selected side is highlighted in red. This routing decision is made before generation begins, and the unselected side does not actually perform any generation tasks.}
\label{A:visual_case}
\end{figure*}

In \cref{A:visual_case}, we demonstrate the visual results of our \texttt{RouteT2I} at a 50\% routing rate, where the router selects the most suitable model between the edge or cloud text-to-image model for each input prompt. The results show that when the edge model performs comparably or even better than the cloud model, the router tends to choose the edge model for generation. Conversely, when the cloud model is significantly superior, it is selected instead. This approach allows us to maintain high generation quality while reducing reliance on cloud services.

\section{Results of More Text-to-Image Model Pairs}\label{other-T2I-comb}

To validate the performance of our routing approach across various edge-cloud T2I model pairs, we conduct experiments not only with SD3 and SD2.1 pairs as shown in Tab. 2 (main text) but also with other model pairs. These T2I models, used on the edge and the cloud, differ in architecture, text encoders, parameter counts, and performance, serving to verify the robustness of our routing method under diverse conditions.

We present the multi-metric image quality performances and their corresponding relative performance improvements for various T2I models on edge and cloud at a routing rate of 50\% in \cref{A:Tab_1} to \cref{A:Tab_17}. Our proposed \texttt{RouteT2I} demonstrates outstanding performance across all combinations. For instance, when SD2.1 is used as the edge model and Pixart-$\alpha$ is used in the cloud in \cref{A:Tab_11}, the performance gap between them is relatively small. In this scenario, \texttt{RouteT2I} achieves its maximum relative performance improvement, enhancing performance by 235.22\% compared to the improvement of cloud models over edge models. Conversely, when SD1.5 is used as the edge model and SD3 is used in the cloud in \cref{A:Tab_3}, the performance gap between them is much larger. In this case, \texttt{RouteT2I} still achieves significant gain, reaching 89.27\% of the improvement brought by the cloud model. Notably, these performance enhancements are achieved at a routing rate of 50\%, meaning that we can reduce cloud model calls by half while still approaching or even surpassing the generation quality of more powerful cloud T2I models. This demonstrates the efficiency of our T2I routing method, which remains effective across diverse combinations of edge and cloud T2I models. By minimizing reliance on costly cloud services, \texttt{RouteT2I} not only reduces operational expenses but also ensures high-quality image generation.

Additionally, the T2I models used as cloud and edge models may have different text encoders. For example, SD1.5 uses CLIP, PixArt-$\alpha$ uses T5, and SD3 uses multiple encoders simultaneously. These encoders map text into different vector spaces, posing challenges for prompt-based routing. However, our method remains effective. For instance, when routing between SDXL and SD1.5 in \cref{A:Tab_9}, which use different numbers of text encoders, our method achieves a quality improvement equivalent to 41.79\% of the cloud model’s gain over the edge model, surpassing other methods by at least 6\%. For routing between models with different encoder types, such as PixArt-$\alpha$ and SD2.1 in \cref{A:Tab_11}, our method reaches an improvement of 235.22\% of the cloud model’s gain, outperforming others by at least 20\%. Notably, for the pair of SD3 and SD1.5 in \cref{A:Tab_3}, where SD3's text encoder far exceed that of SD1.5 in both the number of encoders and parameters, we still achieve an 89.2\% performance gain. This demonstrates that our routing model, equipped with dual-gate token selection MoE, focuses on the semantic meaning of tokens rather than the vector spaces associated with specific encoders. Our approach is consistently efficient across T2I models using diverse text encoders.

We also experiment with the impact of different T2I model architectures on routing. Specifically, we select three mainstream generation architectures: diffusion, diffusion Transformer, and autoregressive. We evaluate routing across different combinations of model types, such as diffusion models, diffusion and diffusion Transformer models, diffusion and autoregressive models, and diffusion Transformer and autoregressive models in \cref{A:Tab_1} to \cref{A:Tab_17}. Our \texttt{RouteT2I} demonstrates consistent effectiveness across all architecture types, achieving significant performance gains in every scenario.



\begin{table*}[]
\centering
\resizebox{\linewidth}{!}{
\begin{tabular}{l|cccccccccc|c}
\toprule
\multirow{2}{*}{Router} & \multicolumn{10}{c|}{Image Quality Metrics}                                                             & \multirow{2}{*}{$\Delta P(\%)$} \\
                        & Definition & Detail & Clarity & Sharpness & Harmony & Realism & Color  & Consistency & Layout & Integrity &                   \\  \midrule
Edge Model: XL-Refiner      & 0.6083          & 0.6584          & 0.5835          & 0.6013          & 0.6086          & 0.5773          & 0.4701          & 0.5195          & 0.4824          & 0.4850          & -               \\
Cloud Model: SD3            & 0.6337          & 0.6847          & 0.6346          & 0.6703          & 0.5930          & 0.5868          & 0.5134          & 0.5199          & 0.5345          & 0.4972          & -               \\ \hdashline
Random                      & 0.6210          & 0.6715          & 0.6091          & 0.6358          & 0.6008          & 0.5821          & 0.4917          & 0.5197          & 0.5084          & 0.4911          & 40.00           \\
RouteLLM-BERT               & 0.6236          & 0.6672          & \textbf{0.6210} & \textbf{0.6431} & 0.6057          & 0.5956          & 0.5024          & 0.5236          & 0.5158          & 0.4963          & 152.48          \\
RouteLLM-MF                 & 0.6231          & 0.6662          & 0.6201          & 0.6429          & 0.6043          & 0.5956          & 0.5022          & 0.5231          & 0.5145          & 0.4955          & 140.40          \\
HybridLLM                   & 0.6239          & \textbf{0.6694} & \textbf{0.6210} & 0.6421          & 0.6044          & 0.5936          & 0.5036          & 0.5237          & \textbf{0.5178} & 0.4951          & 152.13          \\
ZOOTER                      & \textbf{0.6239} & 0.6687          & 0.6186          & 0.6413          & 0.6054          & 0.5945          & 0.5026          & 0.5230          & 0.5168          & 0.4956          & 138.95          \\
\rowcolor[gray]{0.9}
\texttt{RouteT2I} (Ours)    & 0.6231          & 0.6676          & 0.6200          & 0.6410          & \textbf{0.6063} & \textbf{0.5967} & \textbf{0.5039} & \textbf{0.5250} & 0.5164          & \textbf{0.4964} & \textbf{184.92} \\ \bottomrule
\end{tabular}   }
\caption{The multi-dimensional quality of images generated by text-to-image models, with the router selecting between edge and cloud models for each prompt at a 50\% routing rate. The higher the metrics, the better. XL-Refiner is used as the edge model, while SD3 serves as the cloud model.}
\label{A:Tab_1}
\end{table*}

\begin{table*}[]
    \centering
    \resizebox{\linewidth}{!}{
    \begin{tabular}{l|cccccccccc|c}
    \toprule
    \multirow{2}{*}{Router} & \multicolumn{10}{c|}{Image Quality Metrics}                                                             & \multirow{2}{*}{$\Delta P(\%)$} \\
                            & Definition & Detail & Clarity & Sharpness & Harmony & Realism & Color  & Consistency & Layout & Integrity &                   \\  \midrule
Edge Model: SDXL            & 0.5900          & 0.6564          & 0.5645          & 0.5832          & 0.6042          & 0.5660          & 0.4622          & 0.5282          & 0.4866          & 0.4892          & -               \\
Cloud Model: SD3            & 0.6337          & 0.6847          & 0.6346          & 0.6703          & 0.5930          & 0.5868          & 0.5134          & 0.5199          & 0.5345          & 0.4972          & -               \\ \hdashline
Random                      & 0.6119          & 0.6705          & 0.5996          & 0.6267          & 0.5986          & 0.5764          & 0.4878          & 0.5241          & 0.5105          & 0.4932          & 30.00           \\
RouteLLM-BERT               & 0.6161          & 0.6667          & 0.6105          & \textbf{0.6335} & 0.6020          & 0.5893          & 0.4983          & 0.5277          & 0.5171          & 0.4983          & 55.28           \\
RouteLLM-MF                 & 0.6170          & 0.6659          & 0.6103          & 0.6325          & 0.6019          & 0.5896          & 0.4984          & 0.5277          & 0.5172          & 0.4987          & 55.70           \\
HybridLLM                   & 0.6156          & \textbf{0.6694} & 0.6110          & 0.6331          & 0.6012          & 0.5880          & 0.5008          & 0.5257          & \textbf{0.5206} & 0.4978          & 53.08           \\
ZOOTER                      & \textbf{0.6184} & 0.6680          & 0.6107          & 0.6328          & 0.6011          & 0.5882          & 0.5006          & 0.5266          & 0.5198          & 0.4973          & 53.41           \\
\rowcolor[gray]{0.9}
\texttt{RouteT2I} (Ours)    & 0.6155          & 0.6669          & \textbf{0.6121} & 0.6316          & \textbf{0.6031} & \textbf{0.5909} & \textbf{0.5014} & \textbf{0.5286} & 0.5200          & \textbf{0.5002} & \textbf{61.83}  \\\bottomrule
\end{tabular}   }
\caption{The multi-dimensional quality of images generated by text-to-image models, with the router selecting between edge and cloud models for each prompt at a 50\% routing rate. The higher the metrics, the better. SDXL is used as the edge model, while SD3 serves as the cloud model.}
\label{A:Tab_2}
\end{table*}

\begin{table*}[]
    \centering
    \resizebox{\linewidth}{!}{
    \begin{tabular}{l|cccccccccc|c}
    \toprule
    \multirow{2}{*}{Router} & \multicolumn{10}{c|}{Image Quality Metrics}                                                             & \multirow{2}{*}{$\Delta P(\%)$} \\
                            & Definition & Detail & Clarity & Sharpness & Harmony & Realism & Color  & Consistency & Layout & Integrity &                   \\  \midrule
Edge Model: SD1.5           & 0.6035          & 0.6615          & 0.5912          & 0.6454          & 0.5944          & 0.5444          & 0.4659          & 0.5149          & 0.4945          & 0.4730          & -               \\
Cloud Model: SD3            & 0.6337          & 0.6847          & 0.6346          & 0.6703          & 0.5930          & 0.5868          & 0.5134          & 0.5199          & 0.5345          & 0.4972          & -               \\ \hdashline
Random                      & 0.6186          & 0.6731          & 0.6129          & 0.6579          & 0.5937          & 0.5656          & 0.4896          & 0.5174          & 0.5145          & 0.4851          & 40.00           \\
RouteLLM-BERT               & 0.6227          & \textbf{0.6782} & 0.6197          & 0.6635          & 0.5968          & 0.5704          & 0.4947          & 0.5197          & 0.5202          & 0.4861          & 77.48           \\
RouteLLM-MF                 & \textbf{0.6234} & 0.6773          & 0.6195          & 0.6634          & 0.5961          & 0.5708          & 0.4947          & 0.5195          & 0.5199          & 0.4868          & 71.83           \\
HybridLLM                   & 0.6219          & 0.6766          & \textbf{0.6214} & 0.6641          & 0.5970          & 0.5723          & 0.4977          & 0.5199          & 0.5223          & 0.4872          & 80.95           \\
ZOOTER                      & 0.6211          & 0.6762          & 0.6213          & 0.6638          & 0.5966          & 0.5719          & 0.4975          & 0.5197          & 0.5223          & 0.4877          & 76.94           \\
\rowcolor[gray]{0.9}
\texttt{RouteT2I} (Ours)    & 0.6211          & 0.6749          & \textbf{0.6214} & \textbf{0.6643} & \textbf{0.5979} & \textbf{0.5727} & \textbf{0.4985} & \textbf{0.5210} & \textbf{0.5227} & \textbf{0.4880} & \textbf{89.27}  \\\bottomrule
\end{tabular}   }
\caption{The multi-dimensional quality of images generated by text-to-image models, with the router selecting between edge and cloud models for each prompt at a 50\% routing rate. The higher the metrics, the better. SD1.5 is used as the edge model, while SD3 serves as the cloud model.}
\label{A:Tab_3}
\end{table*}

\begin{table*}[]
    \centering
    \resizebox{\linewidth}{!}{
    \begin{tabular}{l|cccccccccc|c}
    \toprule
    \multirow{2}{*}{Router} & \multicolumn{10}{c|}{Image Quality Metrics}                                                             & \multirow{2}{*}{$\Delta P(\%)$} \\
                            & Definition & Detail & Clarity & Sharpness & Harmony & Realism & Color  & Consistency & Layout & Integrity &                   \\  \midrule
Edge Model: XL-Refiner      & 0.6083          & 0.6584          & 0.5835          & 0.6013          & 0.6086          & 0.5773          & 0.4701          & 0.5195          & 0.4824          & 0.4850          & -               \\
Cloud Model: SD2.1          & 0.6251          & 0.6685          & 0.6076          & 0.6537          & 0.5949          & 0.5575          & 0.4680          & 0.5088          & 0.4860          & 0.4690          & -               \\ \hdashline
Random                      & 0.6167          & 0.6634          & 0.5956          & 0.6275          & 0.6018          & 0.5674          & 0.4690          & 0.5141          & 0.4842          & 0.4770          & 0.00            \\
RouteLLM-BERT               & 0.6219          & 0.6649          & 0.6068          & 0.6389          & 0.6065          & 0.5768          & 0.4798          & 0.5180          & 0.4913          & 0.4778          & 94.96           \\
RouteLLM-MF                 & 0.6212          & 0.6638          & 0.6072          & \textbf{0.6391} & 0.6057          & 0.5765          & \textbf{0.4804} & 0.5183          & 0.4913          & 0.4777          & 96.10           \\
HybridLLM                   & 0.6224          & \textbf{0.6653} & 0.6059          & 0.6383          & 0.6072          & 0.5773          & 0.4798          & 0.5184          & 0.4924          & \textbf{0.4781} & 99.58           \\
ZOOTER                      & 0.6220          & \textbf{0.6653} & 0.6052          & 0.6364          & 0.6066          & 0.5766          & 0.4790          & 0.5184          & 0.4910          & 0.4776          & 90.11           \\
\rowcolor[gray]{0.9}
\texttt{RouteT2I} (Ours)    & \textbf{0.6226} & 0.6651          & \textbf{0.6075} & 0.6384          & \textbf{0.6079} & \textbf{0.5778} & \textbf{0.4804} & \textbf{0.5185} & \textbf{0.4926} & 0.4779          & \textbf{104.21} \\\bottomrule
\end{tabular}   }
\caption{The multi-dimensional quality of images generated by text-to-image models, with the router selecting between edge and cloud models for each prompt at a 50\% routing rate. The higher the metrics, the better. XL-Refiner is used as the edge model, while SD2.1 serves as the cloud model.}
\label{A:Tab_4}
\end{table*}

\begin{table*}[]
    \centering
    \resizebox{\linewidth}{!}{
    \begin{tabular}{l|cccccccccc|c}
    \toprule
    \multirow{2}{*}{Router} & \multicolumn{10}{c|}{Image Quality Metrics}                                                             & \multirow{2}{*}{$\Delta P(\%)$} \\
                            & Definition & Detail & Clarity & Sharpness & Harmony & Realism & Color  & Consistency & Layout & Integrity &                   \\  \midrule
Edge Model: SDXL            & 0.5900          & 0.6564          & 0.5645          & 0.5832          & 0.6042          & 0.5660          & 0.4622          & 0.5282          & 0.4866          & 0.4892          & -               \\
Cloud Model: SD2.1          & 0.6251          & 0.6685          & 0.6076          & 0.6537          & 0.5949          & 0.5575          & 0.4680          & 0.5088          & 0.4860          & 0.4690          & -               \\ \hdashline
Random                      & 0.6076          & 0.6624          & 0.5861          & 0.6184          & 0.5996          & 0.5617          & 0.4651          & 0.5185          & 0.4863          & 0.4791          & 0.00            \\
RouteLLM-BERT               & \textbf{0.6142} & 0.6645          & 0.5964          & 0.6290          & 0.6025          & 0.5708          & 0.4751          & 0.5231          & 0.4926          & 0.4799          & 134.47          \\
RouteLLM-MF                 & 0.6141          & 0.6636          & \textbf{0.5967} & \textbf{0.6299} & 0.6019          & 0.5699          & 0.4756          & 0.5231          & 0.4928          & 0.4798          & 136.41          \\
HybridLLM                   & 0.6130          & 0.6642          & 0.5950          & 0.6269          & 0.6039          & 0.5708          & 0.4756          & \textbf{0.5241} & 0.4924          & \textbf{0.4806} & 133.10          \\
ZOOTER                      & \textbf{0.6142} & 0.6641          & 0.5949          & 0.6269          & 0.6041          & 0.5704          & 0.4736          & 0.5227          & 0.4921          & 0.4795          & 124.23          \\
\rowcolor[gray]{0.9}
\texttt{RouteT2I} (Ours)    & 0.6140          & \textbf{0.6646} & 0.5958          & 0.6274          & \textbf{0.6055} & \textbf{0.5715} & \textbf{0.4760} & \textbf{0.5241} & \textbf{0.4934} & 0.4803          & \textbf{152.44} \\\bottomrule
\end{tabular}   }
\caption{The multi-dimensional quality of images generated by text-to-image models, with the router selecting between edge and cloud models for each prompt at a 50\% routing rate. The higher the metrics, the better. SDXL is used as the edge model, while SD2.1 serves as the cloud model.}
\label{A:Tab_5}
\end{table*}

\begin{table*}[]
    \centering
    \resizebox{\linewidth}{!}{
    \begin{tabular}{l|cccccccccc|c}
    \toprule
    \multirow{2}{*}{Router} & \multicolumn{10}{c|}{Image Quality Metrics}                                                             & \multirow{2}{*}{$\Delta P(\%)$} \\
                            & Definition & Detail & Clarity & Sharpness & Harmony & Realism & Color  & Consistency & Layout & Integrity &                   \\  \midrule
Edge Model: SD1.5           & 0.6035          & 0.6615          & 0.5912          & 0.6454          & 0.5944          & 0.5444          & 0.4659          & 0.5149          & 0.4945          & 0.4730          & -               \\
Cloud Model: SD2.1          & 0.6251          & 0.6685          & 0.6076          & 0.6537          & 0.5949          & 0.5575          & 0.4680          & 0.5088          & 0.4860          & 0.4690          & -               \\ \hdashline
Random                      & 0.6143          & 0.6650          & 0.5994          & 0.6496          & 0.5947          & 0.5510          & 0.4669          & 0.5118          & 0.4903          & 0.4710          & 20.00           \\
RouteLLM-BERT               & 0.6183          & 0.6663          & 0.6046          & \textbf{0.6531} & 0.5969          & 0.5570          & 0.4723          & 0.5122          & 0.4926          & 0.4720          & 112.83          \\
RouteLLM-MF                 & 0.6176          & 0.6657          & \textbf{0.6054} & 0.6528          & 0.5959          & 0.5571          & 0.4725          & 0.5125          & 0.4927          & \textbf{0.4726} & 94.83           \\
HybridLLM                   & 0.6167          & 0.6667          & 0.6050          & 0.6523          & \textbf{0.5970} & 0.5556          & 0.4712          & 0.5127          & \textbf{0.4933} & 0.4720          & 109.52          \\
ZOOTER                      & 0.6158          & 0.6662          & 0.6031          & 0.6525          & 0.5969          & 0.5555          & 0.4709          & \textbf{0.5135} & 0.4932          & 0.4715          & 103.70          \\
\rowcolor[gray]{0.9}
\texttt{RouteT2I} (Ours)    & \textbf{0.6186} & \textbf{0.6668} & 0.6048          & 0.6521          & \textbf{0.5970} & \textbf{0.5573} & \textbf{0.4726} & 0.5126          & \textbf{0.4933} & \textbf{0.4726} & \textbf{119.18} \\\bottomrule
\end{tabular}   }
\caption{The multi-dimensional quality of images generated by text-to-image models, with the router selecting between edge and cloud models for each prompt at a 50\% routing rate. The higher the metrics, the better. SD1.5 is used as the edge model, while SD2.1 serves as the cloud model.}
\label{A:Tab_6}
\end{table*}

\begin{table*}[]
    \centering
    \resizebox{\linewidth}{!}{
    \begin{tabular}{l|cccccccccc|c}
    \toprule
    \multirow{2}{*}{Router} & \multicolumn{10}{c|}{Image Quality Metrics}                                                             & \multirow{2}{*}{$\Delta P(\%)$} \\
                            & Definition & Detail & Clarity & Sharpness & Harmony & Realism & Color  & Consistency & Layout & Integrity &                   \\  \midrule
Edge Model: SDXL            & 0.5900          & 0.6564          & 0.5645          & 0.5832          & 0.6042          & 0.5660          & 0.4622          & 0.5282          & 0.4866          & 0.4892          & -               \\
Cloud Model: XL-Refiner     & 0.6083          & 0.6584          & 0.5835          & 0.6013          & 0.6086          & 0.5773          & 0.4701          & 0.5195          & 0.4824          & 0.4850          & -               \\ \hdashline
Random                      & 0.5992          & 0.6574          & 0.5740          & 0.5922          & 0.6064          & 0.5717          & 0.4661          & 0.5239          & 0.4845          & 0.4871          & 20.00           \\
RouteLLM-BERT               & 0.6008          & 0.6567          & 0.5758          & 0.5948          & 0.6108          & 0.5736          & 0.4696          & \textbf{0.5249} & 0.4871          & 0.4874          & 44.04           \\
RouteLLM-MF                 & 0.6012          & 0.6574          & 0.5761          & \textbf{0.5950} & 0.6108          & 0.5742          & 0.4694          & 0.5239          & 0.4870          & 0.4874          & 46.82           \\
HybridLLM                   & 0.6005          & 0.6564          & 0.5759          & 0.5947          & \textbf{0.6115} & 0.5730          & 0.4690          & 0.5245          & 0.4873          & 0.4866          & 40.89           \\
ZOOTER                      & 0.6003          & 0.6574          & 0.5761          & 0.5948          & 0.6110          & 0.5741          & 0.4696          & 0.5245          & 0.4870          & 0.4875          & 48.22           \\
\rowcolor[gray]{0.9}
\texttt{RouteT2I} (Ours)    & \textbf{0.6014} & \textbf{0.6575} & \textbf{0.5763} & \textbf{0.5950} & 0.6113          & \textbf{0.5748} & \textbf{0.4700} & 0.5239          & \textbf{0.4874} & \textbf{0.4876} & \textbf{51.25}  \\\bottomrule
\end{tabular}   }
\caption{The multi-dimensional quality of images generated by text-to-image models, with the router selecting between edge and cloud models for each prompt at a 50\% routing rate. The higher the metrics, the better. SDXL is used as the edge model, while XL-Refiner serves as the cloud model.}
\label{A:Tab_7}
\end{table*}

\begin{table*}[]
    \centering
    \resizebox{\linewidth}{!}{
    \begin{tabular}{l|cccccccccc|c}
    \toprule
    \multirow{2}{*}{Router} & \multicolumn{10}{c|}{Image Quality Metrics}                                                             & \multirow{2}{*}{$\Delta P(\%)$} \\
                            & Definition & Detail & Clarity & Sharpness & Harmony & Realism & Color  & Consistency & Layout & Integrity &                   \\  \midrule
Edge Model: SD1.5           & 0.6035          & 0.6615          & 0.5912          & 0.6454          & 0.5944          & 0.5444          & 0.4659          & 0.5149          & 0.4945          & 0.4730          & -               \\
Cloud Model: XL-Refiner     & 0.6083          & 0.6584          & 0.5835          & 0.6013          & 0.6086          & 0.5773          & 0.4701          & 0.5195          & 0.4824          & 0.4850          & -               \\ \hdashline
Random                      & 0.6059          & 0.6600          & 0.5874          & 0.6234          & 0.6015          & 0.5609          & 0.4680          & 0.5172          & 0.4885          & 0.4790          & 10.00           \\
RouteLLM-BERT               & 0.6090          & 0.6614          & \textbf{0.5983} & 0.6354          & 0.6079          & 0.5693          & 0.4763          & 0.5196          & 0.4945          & 0.4804          & 76.34           \\
RouteLLM-MF                 & 0.6078          & 0.6603          & 0.5977          & \textbf{0.6360} & 0.6075          & 0.5694          & 0.4761          & 0.5198          & 0.4944          & 0.4804          & 69.25           \\
HybridLLM                   & 0.6084          & 0.6606          & 0.5968          & 0.6359          & 0.6073          & 0.5680          & 0.4754          & 0.5191          & 0.4948          & 0.4802          & 66.31           \\
ZOOTER                      & \textbf{0.6104} & \textbf{0.6615} & 0.5964          & 0.6357          & 0.6071          & 0.5677          & 0.4760          & 0.5209          & 0.4951          & 0.4795          & 77.68           \\
\rowcolor[gray]{0.9}
\texttt{RouteT2I} (Ours)    & 0.6091          & \textbf{0.6615} & 0.5969          & 0.6354          & \textbf{0.6082} & \textbf{0.5697} & \textbf{0.4768} & \textbf{0.5212} & \textbf{0.4957} & \textbf{0.4808} & \textbf{80.83}  \\\bottomrule
\end{tabular}   }
\caption{The multi-dimensional quality of images generated by text-to-image models, with the router selecting between edge and cloud models for each prompt at a 50\% routing rate. The higher the metrics, the better. SD1.5 is used as the edge model, while XL-Refiner serves as the cloud model.}
\label{A:Tab_8}
\end{table*}

\begin{table*}[]
    \centering
    \resizebox{\linewidth}{!}{
    \begin{tabular}{l|cccccccccc|c}
    \toprule
    \multirow{2}{*}{Router} & \multicolumn{10}{c|}{Image Quality Metrics}                                                             & \multirow{2}{*}{$\Delta P(\%)$} \\
                            & Definition & Detail & Clarity & Sharpness & Harmony & Realism & Color  & Consistency & Layout & Integrity &                   \\  \midrule
Edge Model: SD1.5            & 0.6035          & 0.6615          & 0.5912          & 0.6454          & 0.5944          & 0.5444          & 0.4659          & 0.5149          & 0.4945          & 0.4730          & -               \\
Cloud Model: SDXL            & 0.5900          & 0.6564          & 0.5645          & 0.5832          & 0.6042          & 0.5660          & 0.4622          & 0.5282          & 0.4866          & 0.4892          & -               \\ \hdashline
Random                       & 0.5968          & 0.6590          & 0.5779          & 0.6143          & 0.5993          & 0.5552          & 0.4640          & 0.5216          & 0.4906          & 0.4811          & -10.00          \\
RouteLLM-BERT                & \textbf{0.6026} & 0.6609          & \textbf{0.5885} & 0.6253          & 0.6019          & 0.5627          & 0.4706          & 0.5232          & 0.4954          & 0.4826          & 35.75           \\
RouteLLM-MF                  & 0.5993          & 0.6602          & 0.5878          & 0.6250          & 0.6034          & 0.5628          & 0.4708          & 0.5241          & 0.4959          & 0.4832          & 35.36           \\
HybridLLM                    & 0.5989          & \textbf{0.6610} & 0.5862          & 0.6237          & 0.6040          & 0.5614          & 0.4700          & 0.5240          & 0.4954          & 0.4833          & 33.08           \\
ZOOTER                       & 0.5992          & 0.6603          & 0.5870          & \textbf{0.6254} & 0.6021          & 0.5608          & 0.4709          & \textbf{0.5248} & 0.4957          & 0.4830          & 33.41           \\
\rowcolor[gray]{0.9}
\texttt{RouteT2I} (Ours)     & 0.6004          & \textbf{0.6610} & 0.5874          & 0.6242          & \textbf{0.6042} & \textbf{0.5635} & \textbf{0.4716} & \textbf{0.5248} & \textbf{0.4960} & \textbf{0.4836} & \textbf{41.79}  \\\bottomrule
\end{tabular}   }
\caption{The multi-dimensional quality of images generated by text-to-image models, with the router selecting between edge and cloud models for each prompt at a 50\% routing rate. The higher the metrics, the better. SD1.5 is used as the edge model, while SDXL serves as the cloud model.}
\label{A:Tab_9}
\end{table*}
 
\begin{table*}[]
    \centering
    \resizebox{\linewidth}{!}{
    \begin{tabular}{l|cccccccccc|c}
    \toprule
    \multirow{2}{*}{Router} & \multicolumn{10}{c|}{Image Quality Metrics}                                                             & \multirow{2}{*}{$\Delta P(\%)$} \\
                            & Definition & Detail & Clarity & Sharpness & Harmony & Realism & Color  & Consistency & Layout & Integrity &                   \\  \midrule
Edge Model: PixArt-$\alpha$  & 0.6660          & 0.7039          & 0.6069          & 0.6499          & 0.6114          & 0.6022          & 0.4562          & 0.4882          & 0.4581          & 0.4753          & -               \\
Cloud Model: SD3             & 0.6337          & 0.6847          & 0.6346          & 0.6703          & 0.5930          & 0.5868          & 0.5134          & 0.5199          & 0.5345          & 0.4972          & -               \\ \hdashline
Random                       & 0.6499          & 0.6943          & 0.6208          & 0.6601          & 0.6022          & 0.5945          & 0.4848          & 0.5041          & 0.4963          & 0.4862          & 10.00           \\
RouteLLM-BERT                & 0.6534          & 0.6917          & 0.6341          & 0.6641          & 0.6062          & 0.6104          & 0.4974          & \textbf{0.5115} & 0.5050          & 0.4911          & 36.93           \\
RouteLLM-MF                  & 0.6545          & 0.6918          & 0.6348          & 0.6654          & 0.6058          & 0.6121          & 0.4970          & 0.5113          & 0.5049          & 0.4927          & 39.71           \\
HybridLLM                    & 0.6527          & \textbf{0.6932} & 0.6339          & 0.6648          & 0.6075          & 0.6119          & \textbf{0.4991} & 0.5105          & \textbf{0.5083} & 0.4938          & 41.10           \\
ZOOTER                       & \textbf{0.6552} & 0.6930          & 0.6344          & 0.6642          & 0.6068          & 0.6124          & 0.4976          & 0.5103          & 0.5064          & 0.4921          & 41.10           \\
\rowcolor[gray]{0.9}
\texttt{RouteT2I} (Ours)     & 0.6543          & 0.6924          & \textbf{0.6353} & \textbf{0.6674} & \textbf{0.6096} & \textbf{0.6136} & 0.4984          & 0.5104          & 0.5081          & \textbf{0.4940} & \textbf{45.13}  \\\bottomrule
\end{tabular}   }
\caption{The multi-dimensional quality of images generated by text-to-image models, with the router selecting between edge and cloud models for each prompt at a 50\% routing rate. The higher the metrics, the better. Pixart-$\alpha$ is used as the edge model, while SD3 serves as the cloud model.}
\label{A:Tab_10}
\end{table*}
 
\begin{table*}[]
    \centering
    \resizebox{\linewidth}{!}{
    \begin{tabular}{l|cccccccccc|c}
    \toprule
    \multirow{2}{*}{Router} & \multicolumn{10}{c|}{Image Quality Metrics}                                                             & \multirow{2}{*}{$\Delta P(\%)$} \\
                            & Definition & Detail & Clarity & Sharpness & Harmony & Realism & Color  & Consistency & Layout & Integrity &                   \\  \midrule
Edge Model: SD2.1            & 0.6251          & 0.6685          & 0.6076          & 0.6537          & 0.5949          & 0.5575          & 0.4680          & 0.5088          & 0.4860          & 0.4690          & -               \\
Cloud Model: PixArt-$\alpha$ & 0.6660          & 0.7039          & 0.6069          & 0.6499          & 0.6114          & 0.6022          & 0.4562          & 0.4882          & 0.4581          & 0.4753          & -               \\ \hdashline
Random                       & 0.6456          & 0.6862          & 0.6073          & 0.6518          & 0.6032          & 0.5798          & 0.4621          & 0.4985          & 0.4720          & 0.4721          & 0.00            \\
RouteLLM-BERT                & 0.6524          & 0.6874          & 0.6183          & 0.6612          & 0.6088          & 0.5926          & 0.4738          & 0.5049          & 0.4812          & 0.4752          & 212.61          \\
RouteLLM-MF                  & 0.6525          & 0.6878          & 0.6175          & 0.6618          & 0.6087          & 0.5921          & 0.4741          & 0.5049          & 0.4813          & 0.4757          & 204.33          \\
HybridLLM                    & \textbf{0.6536} & 0.6880          & 0.6172          & 0.6609          & 0.6084          & 0.5926          & 0.4741          & 0.5058          & 0.4819          & 0.4757          & 197.92          \\
ZOOTER                       & 0.6520          & \textbf{0.6881} & 0.6174          & \textbf{0.6621} & 0.6096          & 0.5916          & 0.4732          & \textbf{0.5064} & 0.4822          & 0.4755          & 203.06          \\
\rowcolor[gray]{0.9}
\texttt{RouteT2I} (Ours)     & 0.6525          & 0.6868          & \textbf{0.6193} & \textbf{0.6621} & \textbf{0.6102} & \textbf{0.5937} & \textbf{0.4757} & 0.5052          & \textbf{0.4831} & \textbf{0.4763} & \textbf{235.22} \\\bottomrule
\end{tabular}   }
\caption{The multi-dimensional quality of images generated by text-to-image models, with the router selecting between edge and cloud models for each prompt at a 50\% routing rate. The higher the metrics, the better. SD2.1 is used as the edge model, while Pixart-$\alpha$ serves as the cloud model.}
\label{A:Tab_11}
\end{table*}
 
\begin{table*}[]
    \centering
    \resizebox{\linewidth}{!}{
    \begin{tabular}{l|cccccccccc|c}
    \toprule
    \multirow{2}{*}{Router} & \multicolumn{10}{c|}{Image Quality Metrics}                                                             & \multirow{2}{*}{$\Delta P(\%)$} \\
                            & Definition & Detail & Clarity & Sharpness & Harmony & Realism & Color  & Consistency & Layout & Integrity &                   \\  \midrule
Edge Model: SDXL             & 0.5900          & 0.6564          & 0.5645          & 0.5832          & 0.6042          & 0.5660          & 0.4622          & 0.5282          & 0.4866          & 0.4892          & -               \\
Cloud Model: PixArt-$\alpha$ & 0.6660          & 0.7039          & 0.6069          & 0.6499          & 0.6114          & 0.6022          & 0.4562          & 0.4882          & 0.4581          & 0.4753          & -              \\ \hdashline
Random                       & 0.6280          & 0.6801          & 0.5857          & 0.6165          & 0.6078          & 0.5841          & 0.4592          & 0.5082          & 0.4724          & 0.4822          & 10.00           \\
RouteLLM-BERT                & 0.6354          & 0.6804          & 0.5940          & 0.6244          & 0.6039          & 0.5933          & 0.4657          & 0.5110          & 0.4729          & 0.4809          & 22.09           \\
RouteLLM-MF                  & 0.6354          & 0.6807          & 0.5977          & \textbf{0.6268} & 0.6076          & 0.5962          & 0.4685          & 0.5125          & 0.4792          & 0.4827          & 37.97           \\
HybridLLM                    & 0.6356          & \textbf{0.6814} & 0.5962          & 0.6239          & 0.6093          & 0.5956          & 0.4675          & 0.5120          & 0.4793          & 0.4828          & 37.81           \\
ZOOTER                       & 0.6350          & 0.6804          & 0.5942          & 0.6241          & \textbf{0.6116} & 0.5940          & 0.4668          & 0.5115          & 0.4787          & 0.4836          & 38.84           \\
\rowcolor[gray]{0.9}
\texttt{RouteT2I} (Ours)     & \textbf{0.6360} & 0.6805          & \textbf{0.5981} & 0.6256          & 0.6101          & \textbf{0.5973} & \textbf{0.4695} & \textbf{0.5130} & \textbf{0.4801} & \textbf{0.4837} & \textbf{44.34}  \\\bottomrule
\end{tabular}   }
\caption{The multi-dimensional quality of images generated by text-to-image models, with the router selecting between edge and cloud models for each prompt at a 50\% routing rate. The higher the metrics, the better. SDXL is used as the edge model, while Pixart-$\alpha$ serves as the cloud model.}
\label{A:Tab_12}
\end{table*}
 
\begin{table*}[]
    \centering
    \resizebox{\linewidth}{!}{
    \begin{tabular}{l|cccccccccc|c}
    \toprule
    \multirow{2}{*}{Router} & \multicolumn{10}{c|}{Image Quality Metrics}                                                             & \multirow{2}{*}{$\Delta P(\%)$} \\
                            & Definition & Detail & Clarity & Sharpness & Harmony & Realism & Color  & Consistency & Layout & Integrity &                   \\  \midrule
Edge Model: XL-Refiner       & 0.6083          & 0.6584          & 0.5835          & 0.6013          & 0.6086          & 0.5773          & 0.4701          & 0.5195          & 0.4824          & 0.4850          & -               \\
Cloud Model: PixArt-$\alpha$ & 0.6660          & 0.7039          & 0.6069          & 0.6499          & 0.6114          & 0.6022          & 0.4562          & 0.4882          & 0.4581          & 0.4753          & -               \\ \hdashline
Random                       & 0.6371          & 0.6811          & 0.5952          & 0.6256          & 0.6100          & 0.5898          & 0.4632          & 0.5039          & 0.4702          & 0.4801          & 10.00           \\
RouteLLM-BERT                & 0.6454          & 0.6812          & \textbf{0.6075} & 0.6341          & 0.6149          & 0.6031          & 0.4738          & \textbf{0.5099} & 0.4793          & 0.4813          & 55.85           \\
RouteLLM-MF                  & 0.6453          & 0.6806          & 0.6062          & 0.6342          & 0.6131          & 0.6031          & 0.4738          & 0.5097          & 0.4793          & 0.4809          & 48.20           \\
HybridLLM                    & \textbf{0.6458} & 0.6805          & 0.6063          & \textbf{0.6345} & 0.6132          & 0.6034          & 0.4737          & \textbf{0.5099} & 0.4784          & 0.4814          & 48.98           \\
ZOOTER                       & 0.6449          & \textbf{0.6814} & 0.6054          & \textbf{0.6345} & 0.6138          & 0.6007          & 0.4719          & 0.5094          & 0.4781          & 0.4823          & 48.84           \\
\rowcolor[gray]{0.9}
\texttt{RouteT2I} (Ours)     & 0.6434          & 0.6793          & 0.6058          & 0.6328          & \textbf{0.6165} & \textbf{0.6040} & \textbf{0.4749} & 0.5093          & \textbf{0.4799} & \textbf{0.4830} & \textbf{62.41}  \\\bottomrule
\end{tabular}   }
\caption{The multi-dimensional quality of images generated by text-to-image models, with the router selecting between edge and cloud models for each prompt at a 50\% routing rate. The higher the metrics, the better. XL-Refiner is used as the edge model, while Pixart-$\alpha$ serves as the cloud model.}
\label{A:Tab_13}
\end{table*}
 
\begin{table*}[]
    \centering
    \resizebox{\linewidth}{!}{
    \begin{tabular}{l|cccccccccc|c}
    \toprule
    \multirow{2}{*}{Router} & \multicolumn{10}{c|}{Image Quality Metrics}                                                             & \multirow{2}{*}{$\Delta P(\%)$} \\
                            & Definition & Detail & Clarity & Sharpness & Harmony & Realism & Color  & Consistency & Layout & Integrity &                   \\  \midrule
Edge Model: SD1.5            & 0.6035          & 0.6615          & 0.5912          & 0.6454          & 0.5944          & 0.5444          & 0.4659          & 0.5149          & 0.4945          & 0.4730          & -               \\
Cloud Model: PixArt-$\alpha$ & 0.6660          & 0.7039          & 0.6069          & 0.6499          & 0.6114          & 0.6022          & 0.4562          & 0.4882          & 0.4581          & 0.4753          & -               \\ \hdashline
Random                       & 0.6347          & 0.6827          & 0.5991          & 0.6477          & 0.6029          & 0.5733          & 0.4611          & 0.5016          & 0.4763          & 0.4741          & 20.00           \\
RouteLLM-BERT                & 0.6382          & \textbf{0.6844} & 0.6094          & 0.6580          & 0.6082          & 0.5840          & 0.4701          & \textbf{0.5100} & 0.4830          & 0.4781          & 87.13           \\
RouteLLM-MF                  & 0.6384          & 0.6837          & 0.6093          & 0.6588          & 0.6106          & 0.5854          & 0.4710          & 0.5095          & 0.4849          & 0.4776          & 89.28           \\
HybridLLM                    & 0.6395          & 0.6837          & 0.6085          & 0.6573          & 0.6107          & \textbf{0.5861} & 0.4717          & 0.5091          & 0.4856          & 0.4780          & 88.25           \\
ZOOTER                       & \textbf{0.6400} & 0.6839          & 0.6086          & 0.6587          & \textbf{0.6108} & 0.5842          & 0.4711          & 0.5075          & 0.4860          & 0.4781          & 90.87           \\
\rowcolor[gray]{0.9}
\texttt{RouteT2I} (Ours)     & 0.6393          & 0.6830          & \textbf{0.6104} & \textbf{0.6589} & \textbf{0.6108} & 0.5859          & \textbf{0.4723} & 0.5068          & \textbf{0.4861} & \textbf{0.4785} & \textbf{95.00}  \\\bottomrule
\end{tabular}   }
\caption{The multi-dimensional quality of images generated by text-to-image models, with the router selecting between edge and cloud models for each prompt at a 50\% routing rate. The higher the metrics, the better. SD1.5 is used as the edge model, while Pixart-$\alpha$ serves as the cloud model.}
\label{A:Tab_14}
\end{table*}
 
\begin{table*}[]
    \centering
    \resizebox{\linewidth}{!}{
    \begin{tabular}{l|cccccccccc|c}
    \toprule
    \multirow{2}{*}{Router} & \multicolumn{10}{c|}{Image Quality Metrics}                                                             & \multirow{2}{*}{$\Delta P(\%)$} \\
                            & Definition & Detail & Clarity & Sharpness & Harmony & Realism & Color  & Consistency & Layout & Integrity &                   \\  \midrule
Edge Model: Infinity         & 0.5329          & 0.6810          & 0.5577          & 0.5817          & 0.6222          & 0.5817          & 0.4587          & 0.4926          & 0.4766          & 0.5009          & -               \\
Cloud Model: SD3             & 0.6337          & 0.6847          & 0.6346          & 0.6703          & 0.5930          & 0.5868          & 0.5134          & 0.5199          & 0.5345          & 0.4972          & -               \\ \hdashline
Random                       & 0.5833          & 0.6828          & 0.5961          & 0.6260          & 0.6076          & 0.5843          & 0.4860          & 0.5063          & 0.5055          & 0.4990          & 30.00           \\
RouteLLM-BERT                & 0.5879          & \textbf{0.6866} & 0.6031          & \textbf{0.6413} & 0.6154          & 0.5947          & 0.4929          & \textbf{0.5140} & 0.5117          & 0.5017          & 78.93           \\
RouteLLM-MF                  & 0.5899          & 0.6849          & 0.6036          & \textbf{0.6413} & 0.6156          & 0.5962          & 0.4955          & 0.5135          & 0.5135          & 0.5026          & 80.57           \\
HybridLLM                    & 0.5932          & 0.6837          & 0.6093          & 0.6370          & 0.6196          & 0.6028          & \textbf{0.5031} & 0.5074          & 0.5210          & 0.5035          & 95.28           \\
ZOOTER                       & 0.5941          & 0.6853          & 0.6067          & 0.6365          & 0.6211          & 0.6030          & 0.5000          & 0.5087          & 0.5198          & 0.5025          & 97.54           \\
\rowcolor[gray]{0.9}
\texttt{RouteT2I} (Ours)     & \textbf{0.5947} & 0.6854          & \textbf{0.6095} & 0.6384          & \textbf{0.6218} & \textbf{0.6056} & 0.5018          & 0.5095          & \textbf{0.5212} & \textbf{0.5036} & \textbf{106.87} \\\bottomrule
\end{tabular}   }
\caption{The multi-dimensional quality of images generated by text-to-image models, with the router selecting between edge and cloud models for each prompt at a 50\% routing rate. The higher the metrics, the better. Infinity is used as the edge model, while SD3 serves as the cloud model.}
\label{A:Tab_15}
\end{table*}
 
\begin{table*}[]
    \centering
    \resizebox{\linewidth}{!}{
    \begin{tabular}{l|cccccccccc|c}
    \toprule
    \multirow{2}{*}{Router} & \multicolumn{10}{c|}{Image Quality Metrics}                                                             & \multirow{2}{*}{$\Delta P(\%)$} \\
                            & Definition & Detail & Clarity & Sharpness & Harmony & Realism & Color  & Consistency & Layout & Integrity &                   \\  \midrule
Edge Model: Infinity         & 0.5329          & 0.6810          & 0.5577          & 0.5817          & 0.6222          & 0.5817          & 0.4587          & 0.4926          & 0.4766          & 0.5009          & -               \\
Cloud Model: SD2.1           & 0.6251          & 0.6685          & 0.6076          & 0.6537          & 0.5949          & 0.5575          & 0.4680          & 0.5088          & 0.4860          & 0.4690          & -               \\ \hdashline
Random                       & 0.5790          & 0.6747          & 0.5827          & 0.6177          & 0.6086          & 0.5696          & 0.4633          & 0.5007          & 0.4813          & 0.4850          & 10.00           \\
RouteLLM-BERT                & 0.5842          & \textbf{0.6774} & 0.5960          & 0.6313          & 0.6246          & 0.5872          & 0.4777          & 0.5043          & 0.4978          & \textbf{0.4922} & 67.89           \\
RouteLLM-MF                  & 0.5812          & 0.6758          & 0.5958          & \textbf{0.6334} & 0.6219          & 0.5839          & 0.4765          & \textbf{0.5047} & 0.4940          & 0.4921          & 59.12           \\
HybridLLM                    & 0.5888          & 0.6755          & 0.5946          & 0.6317          & 0.6238          & 0.5869          & 0.4774          & 0.5034          & 0.4956          & 0.4880          & 61.66           \\
ZOOTER                       & 0.5885          & 0.6764          & 0.5937          & 0.6314          & 0.6237          & 0.5861          & 0.4767          & 0.5045          & 0.4955          & 0.4884          & 61.68           \\
\rowcolor[gray]{0.9}
\texttt{RouteT2I} (Ours)     & \textbf{0.5889} & 0.6760          & \textbf{0.5980} & 0.6332          & \textbf{0.6279} & \textbf{0.5909} & \textbf{0.4803} & 0.5040          & \textbf{0.5001} & 0.4899          & \textbf{75.02}  \\\bottomrule
\end{tabular}   }
\caption{The multi-dimensional quality of images generated by text-to-image models, with the router selecting between edge and cloud models for each prompt at a 50\% routing rate. The higher the metrics, the better. Infinity is used as the edge model, while SD2.1 serves as the cloud model.}
\label{A:Tab_16}
\end{table*}
 
\begin{table*}[]
    \centering
    \resizebox{\linewidth}{!}{
    \begin{tabular}{l|cccccccccc|c}
    \toprule
    \multirow{2}{*}{Router} & \multicolumn{10}{c|}{Image Quality Metrics}                                                             & \multirow{2}{*}{$\Delta P(\%)$} \\
                            & Definition & Detail & Clarity & Sharpness & Harmony & Realism & Color  & Consistency & Layout & Integrity &                   \\  \midrule
Edge Model: Infinity         & 0.5329          & 0.6810          & 0.5577          & 0.5817          & 0.6222          & 0.5817          & 0.4587          & 0.4926          & 0.4766          & 0.5009          & -               \\
Cloud Model: PixArt-$\alpha$ & 0.6660          & 0.7039          & 0.6069          & 0.6499          & 0.6114          & 0.6022          & 0.4562          & 0.4882          & 0.4581          & 0.4753          & -                \\ \hdashline
Random                       & 0.5995          & 0.6924          & 0.5823          & 0.6158          & 0.6168          & 0.5920          & 0.4574          & 0.4904          & 0.4673          & 0.4881          & 0.00            \\
RouteLLM-BERT                & 0.6104          & \textbf{0.6924} & 0.5981          & \textbf{0.6333} & 0.6309          & 0.6185          & 0.4729          & 0.5002          & 0.4824          & 0.4926          & 127.82          \\
RouteLLM-MF                  & 0.6064          & 0.6909          & 0.5962          & 0.6329          & 0.6259          & 0.6162          & 0.4711          & \textbf{0.5013} & 0.4782          & 0.4928          & 113.45          \\
HybridLLM                    & 0.6088          & 0.6921          & 0.5979          & 0.6311          & 0.6304          & 0.6159          & 0.4735          & 0.4981          & 0.4827          & 0.4932          & 123.07          \\
ZOOTER                       & \textbf{0.6105} & \textbf{0.6924} & 0.5960          & 0.6312          & 0.6309          & 0.6169          & 0.4718          & 0.4985          & 0.4822          & 0.4918          & 117.25          \\
\rowcolor[gray]{0.9}
\texttt{RouteT2I} (Ours)     & 0.6083          & 0.6910          & \textbf{0.5983} & 0.6294          & \textbf{0.6327} & \textbf{0.6187} & \textbf{0.4743} & 0.4978          & \textbf{0.4853} & \textbf{0.4937} & \textbf{130.24} \\ \bottomrule
\end{tabular}   }
\caption{The multi-dimensional quality of images generated by text-to-image models, with the router selecting between edge and cloud models for each prompt at a 50\% routing rate. The higher the metrics, the better. Infinity is used as the edge model, while Pixart-$\alpha$ serves as the cloud model.}
\label{A:Tab_17}
\end{table*}

\end{document}